\documentclass{article}
\usepackage{graphicx} 
\usepackage[left]{lineno} 
\usepackage[a4paper, margin=1in]{geometry} 
\usepackage{breqn}
\usepackage{breqn}
\usepackage{xcolor}
\usepackage{amssymb}     
\usepackage{amsfonts}    
\usepackage{mathtools}   
\usepackage{bm}          
\usepackage{algorithm}        
\usepackage{algorithmic}      
\usepackage{listings}         
\usepackage{booktabs}    
\usepackage{array}       
\usepackage{multirow}    
\usepackage{tabularx}    
\usepackage{textcomp}

\usepackage{graphicx} 
\usepackage[left]{lineno} 
\usepackage[a4paper, margin=1in]{geometry} 
\usepackage{amsmath}     
\usepackage{amssymb}     
\usepackage{amsfonts}    
\usepackage{mathtools}   
\usepackage{bm}          
\usepackage{algorithm}        
\usepackage{algorithmic}      
\usepackage{listings}         
\usepackage{booktabs}    
\usepackage{array}       
\usepackage{multirow}    
\usepackage{tabularx}    
\usepackage{setspace}    

\usepackage[style=nature, sorting=none, backend=bibtex]{biblatex}
\date{} 

\begin{document}

\begin{flushleft}
    \Large \textbf{Multi-resolution Score-Based Variational Graphical Diffusion for Causal Disaster System Modeling and Inference} \\
    \vspace{0.5cm}
    \normalsize
    Xuechun Li\textsuperscript{1}, Shan Gao\textsuperscript{1}, Susu Xu\textsuperscript{1,2}\\
    \textsuperscript{1}\textit{Center for Systems Science and Engineering, Department of Civil and Systems Engineering, Johns Hopkins University}\\
    \textsuperscript{2}\textit{Data Science and AI Institute, Johns Hopkins University}\\
\end{flushleft}

\noindent \textbf{Complex systems with intricate causal dependencies present significant challenges for accurate estimation and prediction. Effective modeling of these systems requires precise modeling of the underlying physical processes, integration of multiple interdependent factors, and incorporation of multi-source observational data with varying resolutions. These systems can manifest in both static scenarios with instantaneous causal chains and temporal scenarios with evolving dynamics, further complicating the modeling challenge. Existing methods struggle to simultaneously handle varying resolutions, capture physical relationships, model causal dependencies, and incorporate temporal dynamics, especially when data comes from diverse sources with inconsistent sampling. Therefore, we introduce Temporal-SVGDM: Score-based Variational Graphical Diffusion Model for Multi-resolution observations, which addresses these challenges through a novel integration of score-based diffusion models and causal graphical models. Our framework first constructs individual SDEs for each variable at its corresponding native resolution, then couples these SDEs through a causal score mechanism where parent nodes inform the evolution of the child nodes. This approach enables unified modeling of both immediate causal effects in static scenarios and evolving dependencies in temporal scenarios. In temporal models, these state representations are then processed through a temporal sequence prediction model to predict future states based on historical patterns and causal relationships. Through extensive experiments on real-world datasets, we demonstrate not only improved prediction accuracy and causal understanding compared to existing methods, but also robust performance under varying levels of background knowledge availability. Our model exhibits graceful degradation patterns across different disaster types, successfully handling both static earthquake scenarios and temporal hurricane and wildfire scenarios, while maintaining superior performance even with limited data.}

\section{Introduction}

Complex systems with intricate causal dependencies, such as those in disaster modeling \cite{cutter2015global} and climate science \cite{palmer2019scientific}, present significant challenges to accurate estimation of the underlying complex systems. These systems can manifest in both static and temporal contexts. In earthquakes, for instance, the causal chain operates almost instantaneously - ground shaking directly causes building damage while simultaneously triggering multiple cascading hazards, such as landslides and land liquefaction, all of which contribute to the final damage state. In contrast, wildfires unfold as a temporal sequence, where initial ignition combines with multiple interconnected dynamic factors, for example, a small forest fire that begins in calm conditions can rapidly intensify when afternoon winds shift and humidity drops. This progression depends on the temporally evolving environmental conditions, including fuel conditions, weather conditions, and topography. These diverse scenarios demonstrate how natural disasters can exhibit both immediate and static causal relationships and temporally evolving dependencies. Accurate modeling of such systems requires three critical elements \cite{xu2022seismic,li2025rapid,li2023disasternet,wang2023causality}: 1) accurate representation of the underlying physical processes in both static and temporal contexts, 2) integration of multiple factors with their immediate and evolving causal relationships, and 3) incorporation of observational data and existing physical knowledge across various spatial and temporal scales.

However, achieving these goals is complicated by several fundamental challenges in modeling complex systems. First, capturing the underlying physical processes requires sophisticated modeling techniques that can handle both non-linear interactions and temporal evolution across multiple scales. Second, integrating causal dependencies demands a framework that can represent and learn complex hierarchical relationships among variables across both space and time, while preserving the physical meaning of these relationships. Third, the nature of available data poses unique challenges: background knowledge relevant to these systems often comes from diverse sources characterized by noise, sparseness, and varying resolutions, where resolution refers to the level of detail in the data, with temporal data often having inconsistent sampling rates and missing time points.

Despite the surge in interest, modeling complex systems with varying spatial and temporal scales remains challenging. Simple interpolation approaches that only rely on upsampling/downsampling techniques \cite{de2011multi, xu2014assimilation} often lead to information loss, artificial detail creation, and misrepresentation of scale-dependent phenomena \cite{atkinson2013downscaling}. While advanced deep learning methods for resolution fusion \cite{song2019survey} attempt to address these limitations, they often fail to capture the underlying physical relationships.
More recent probabilistic graphical models \cite{xu2022deep, xu2022seismic, li2023normalizing, li2024multi, li2024optimizing, li2024scalable, li2024rapid, li2024spatial} have shown promise in modeling complex causal dependencies among observed and unobserved variables. However, these approaches make a critical assumption of consistent observational resolution and coverage across all variables, making them unsuitable for real-world scenarios where background knowledge come at varying resolutions. Moreover, these models focus primarily on static causal relationships, overlooking the crucial temporal aspects of disaster evolution.
The temporal modeling community has developed sophisticated approaches like ConvLSTM \cite{shi2015convolutional} for sequence learning and attention mechanisms \cite{vaswani2017attention} for capturing long-range dependencies. However, while these models excel at pattern recognition in time series data, they lack the ability to capture and represent explicit causal relationships between variables, limiting their applicability in physically-driven systems. Even recent attempts at unifying spatial and temporal modeling through neural ODEs \cite{chen2018neural} and physics-informed neural networks \cite{raissi2019physics} face fundamental limitations: they require uniform sampling in both space and time, a condition rarely met in real-world disaster monitoring where data streams come from diverse sources with varying temporal and spatial resolutions.

Recently, diffusion models \cite{sohl2015deep} have gained traction in various fields due to their ability to generate high-quality samples and handle complex data distributions. They offer advantages over other generative models like generative adversarial networks (GANs) and variational autoencoders (VAEs) in stability, sample quality, and diversity \cite{arjovsky2017wasserstein, ho2020denoising, kingma2021variational, dhariwal2021diffusion}. Particularly, score-based diffusion models offer several advantages \cite{song2020improved}. Unlike Denoising Diffusion Probabilistic Models (DDPM) \cite{ho2020denoising}, which focus on learning the reverse process of a fixed forward diffusion process, score-based models directly learn the score function of the data distribution. This approach allows for more flexible noise schedules and can potentially lead to faster sampling and better quality results, especially in complex, high-dimensional spaces typical of earth system data \cite{song2021maximum}. Moreover, score-based stochastic differential equation (SDE) models are effective for handling stochasticity, which better align with the evolving and unpredictable nature of earth systems, and offer potential advantages in interpretability by directly modeling the gradient of the log-density of the data distribution and providing a continuous representation that can align with physical processes. This approach can offer insights into the local structures of complex data distributions typical in earth system, potentially aiding in the understanding of underlying phenomena in causal disaster systems.

In this work, we introduce Temporal Score-based Variational Graphical Diffusion Model (Temporal-SVGDM), a framework designed to handle both the spatial and temporal aspects of complex systems with multi-resolution background knowledge and causal dependencies. In causal disaster systems, multi-resolution background knowledge corresponds to different latent variables in our framework. For each latent variable, there may be zero to multiple pieces of corresponding background knowledge at different spatial resolutions, depending on data availability and measurement techniques\cite{li2025multi}. High-resolution background knowledge captures fine-scale details but may be sensitive to local anomalies, while low-resolution data provide broader patterns at the expense of local precision. Each resolution offers unique insights into the underlying process. Our framework recognizes that many real-world processes are continuous in nature, with both immediate causal effects and temporal evolution, while our background knowledge is inherently discrete and limited by the resolution of our measurement tools. The framework aims to integrate background knowledge across different resolutions to approximate the underlying continuous process as closely as possible. To clarify, multi-resolution background knowledge in our context refer to data collected at varying spatial scales for the same or different phenomena in a system. Temporal-SVGDM integrates these diverse resolutions, balancing detailed local information with broader regional patterns and temporal evolution to comprehensively model the complex, causal disaster system. 

\begin{figure}[htbp]
  \centering
\includegraphics[width=1.1\linewidth]{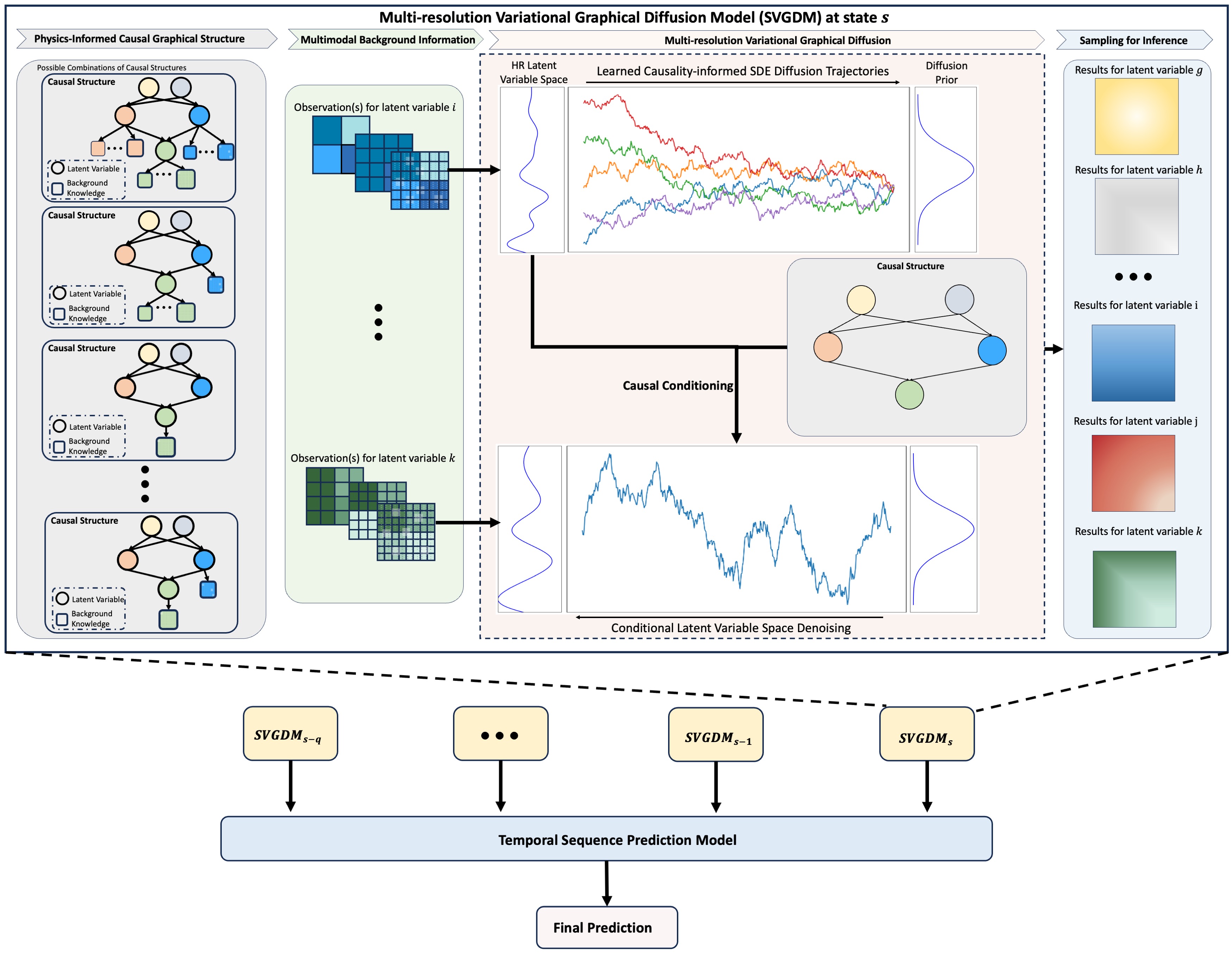}
\vspace{-0.8cm}
  \caption{\textbf{Overview of the Temporal Multi-resolution Variational Graphical Diffusion Model (Temporal-SVGDM).} At each time step, the framework consists of: (1) A physics-informed causal graphical structure that supports multiple possible causal relationships between latent variables and multi-resolution background knowledge, integrating them while distinctly processing parent and child nodes, (2) A multi-modal background information module that processes background knowledge at their native resolutions while respecting causal hierarchies, (3) A multi-resolution variational graphical diffusion component where parent node SDE values are used to compute causal scores, which combine with unconditional scores to enable causality-informed diffusion trajectories and conditional latent variable space denoising, and (4) Sampling for inference to generate results for latent variables. The outputs from multiple time steps are then fed into a temporal sequence prediction model to generate final predictions, enabling the framework to capture both spatial and temporal dependencies in complex physical systems.}
  \vspace{-0.8cm}
  \label{Overview}
  \end{figure}

To jointly model all factors with their causal dependency in a complex system, we formulate these systems together with their multi-resolution background knowledge as a new family of Bayesian networks (BNs). Temporal-SVGDM first constructs individual stochastic differential equations (SDEs) for each variable at its corresponding native resolution, preserving the intrinsic characteristics of the background knowledge. These SDEs are then coupled through a causal score mechanism where outputs from parent node SDEs inform the causal score of their child nodes. This approach enables our framework to maintain the fidelity of individual processes while facilitating meaningful causal interactions across different resolutions. The framework then processes these state representations through a temporal sequence prediction model, which learns to predict future states based on historical patterns and causal relationships. By incorporating both spatial and temporal dynamics, and by maintaining causal integrity throughout the modeling process, our framework offers a powerful tool for modeling causal disaster systems. The overview of Temporal-SVGDM is shown in Figure \ref{Overview}. We demonstrate the effectiveness of our approach through experiments on real-world disasters designed to mimic the complexity of disaster-induced cascading hazards processes, showing improved performance in both prediction accuracy and causal understanding. Our results demonstrate the ability of Temporal-SVGDM to not only model spatial relationships but also accurately predict temporal evolution of complex systems. 

Through extensive experiments on real-world disasters, we demonstrate not only the effectiveness of our approach in terms of prediction accuracy and causal understanding but also its robustness to varying levels of background knowledge availability. Our model maintains superior performance compared to existing methods even under limited data conditions, with graceful degradation patterns that differ meaningfully across disaster types - from relatively modest drops in earthquake assessment (4-5\% reduction in AUC) to more pronounced but still competitive performance changes in hurricane (12.32\% AUC reduction) and wildfire (13\% F1 score reduction) prediction. These results highlight the practical utility of our framework in real-world scenarios where data availability and quality often vary significantly. The ability of Temporal-SVGDM to not only model spatial relationships but also accurately predict temporal evolution while maintaining robust performance under different background knowledge conditions demonstrates its potential as a comprehensive solution for complex system modeling. The potential applications of our framework can be extended beyond disaster modeling to various environmental and geoscience domains.

\section{Results}
\subsection{Causal Bayesian network for modeling complex disaster systems}

\begin{figure*}[ht]
  \centering
\includegraphics[width=1\linewidth]{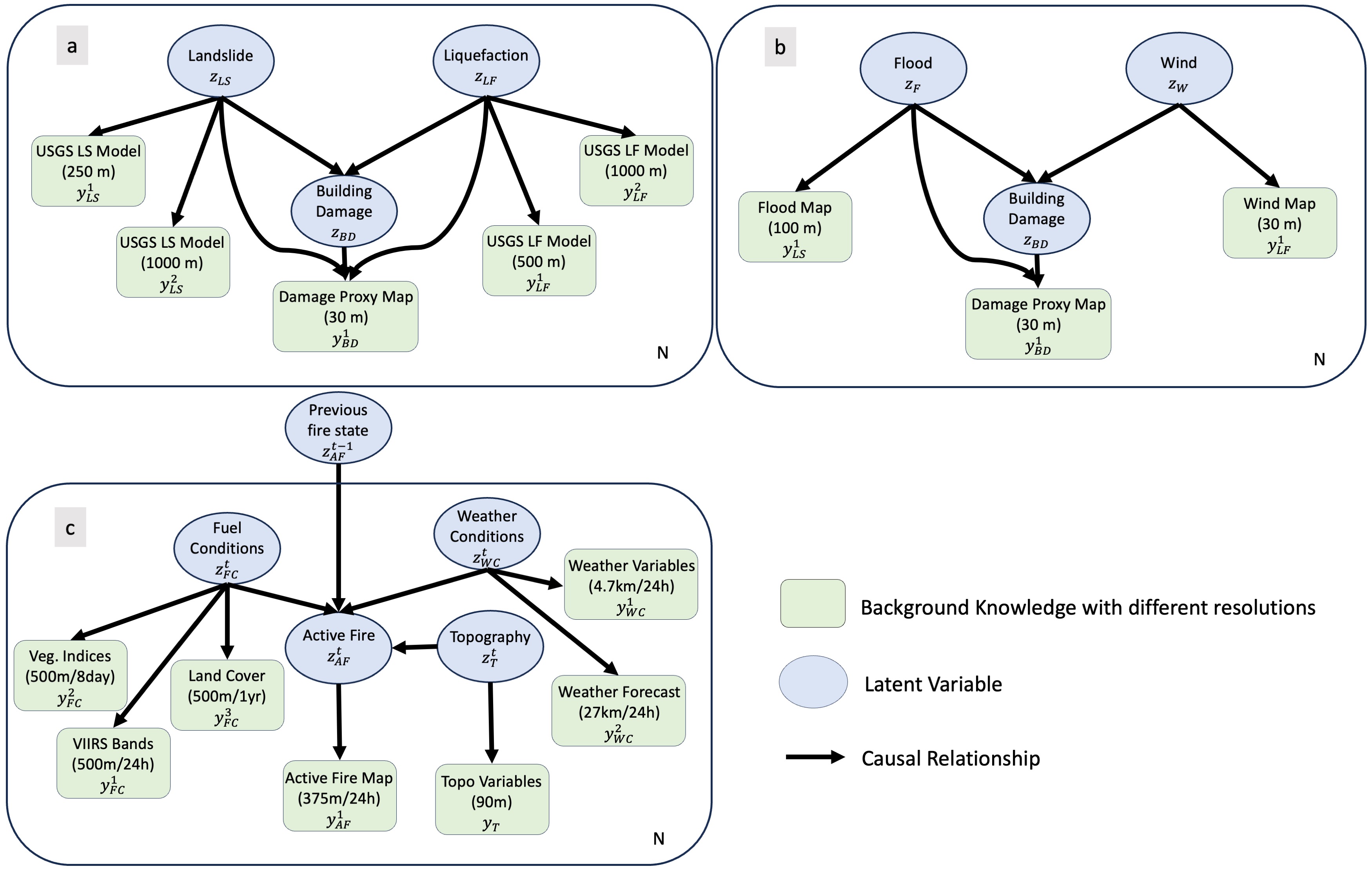}
  \caption{\textbf{Examples of our causal Bayesian inference framework with multi-resolution background knowledge for causal disaster systems.} Figure (a) shows the overview of causal Bayesian inference framework for seismic multi-hazard and impacts estimation. Figure (b) presents the causal Bayesian network for hurricane-induced building damage estimation. Figure (c) shows the causal Bayesian network for fire spread estimation.}
  \label{BNs}
    \end{figure*}

We design causal Bayesian networks to capture the complex dependencies in different disaster systems while incorporating multi-resolution background knowledge. These networks encode our assumptions about both the causal structure of disaster processes and how they relate to available observational data. In our framework, each network consists of two fundamental types of nodes. The first type, latent variables (shown in blue circle in Figure \ref{BNs}), represents the underlying physical states we aim to estimate. The second type, background knowledge (shown in green rectangle), provides observable information at various resolutions. These nodes represent different types of information that inform our understanding of the latent states. Background knowledge can come at different resolutions due to varying measurement techniques, data sources, and modeling approaches.

The relationship between these variables follows a hierarchical structure, where latent variables form the core causal network while background knowledge serves as evidence for inferring these latent states. We use a systematic notation system where $z$ represents latent variables with subscripts indicating their physical meaning (e.g., $z_{LS}$ for landslide, $z_{LF}$ for liquefaction), and $y$ represents background knowledge with subscripts denoting both the corresponding physical process and the specific data source. When temporal relationships are involved, we use superscripts to indicate temporal indices (e.g., $z^{t-1}$ for previous states).

The seismic hazard network (Figure \ref{BNs} (a)) demonstrates a complex cascading structure where earthquake-induced hazards, such as landslide ($z_{LS}$) and liquefaction ($z_{LF}$), combine to cause building damage ($z_{BD}$). Each latent variable is informed by background knowledge at different resolutions. For example, landslide susceptibility is informed by USGS Landslide Models at both 250-meter (high resolution) and 1000-meter (low resolution) resolutions. This multi-resolution approach allows us to capture both fine-grained local features and broader regional patterns.

The hurricane scenario (Figure \ref{BNs}(b)) illustrates how our framework handles concurrent hazards. The network models how flood conditions ($z_{F}$) and wind forces ($z_{W}$) simultaneously affect building damage ($z_{BD}$). Both flood and wind represent distinct physical processes that can independently cause structural damage. The background knowledge spans multiple resolutions: flood maps at 100-meter resolution, wind field data at 30-meter, and damage proxy maps at 30-meter, demonstrating the ability of our framework to integrate heterogeneous data sources. This diverse set of background knowledge demonstrates the ability of our framework to integrate heterogeneous data sources while preserving their native resolutions.

The fire spread network (Figure \ref{BNs}(c)) showcases the temporal aspects of our framework through the explicit modeling of previous fire states ($z_{t-1}^{F}$). The network captures complex interactions between fuel conditions ($z_{FC}$), weather conditions ($z_{WC}$), and topography ($z_{T}$), each informed by multiple data sources at varying resolutions. Each of these factors is informed by multiple data sources at varying resolutions - vegetation indices from satellite observations provide fuel condition information at different temporal frequencies, weather stations offer meteorological measurements at various spatial densities, and digital elevation models contribute topographical data at different scales. The temporal dependencies between these variables, particularly through the previous fire state, allow our model to capture both the immediate and evolving dynamics of wildfire progression.

\begin{table*}[htbp]
\begin{center}
\caption{This table presents the comparison results (AUC) using real-world seismic data with baseline models for three unobserved variables (landslide, liquefaction, and building damage). NGA means no ground truth label is available. - means no results exist.}
\scalebox{1}{
\begin{tabular}{c|c|cccc}
\toprule
\textbf{Earthquake} & \textbf{Model}   & \textbf{$z_{\text{LS}}$} & \textbf{$z_{\text{LF}}$} & \textbf{$z_{\text{BD}}$}\\
\hline
 \multirow{6}{*}{2020 Puerto Rico earthquake} & \textbf{SVGDM} & \textbf{0.9581} & \textbf{0.9517} & \textbf{0.9612}
\\
& VCBI \cite{xu2022seismic} & 0.9012 & 0.9034 & 0.9123
 \\ 
 & DisasterNet \cite{li2023disasternet} & 0.9293 & 0.9284 & 0.9413\\
   & Prior Model \cite{zhu2015geospatial, nowicki2018global}  & 0.8712 & 0.8913 & -\\
&  BBVI \cite{ranganath2014black} & 0.7912 & 0.7731 & 0.7423
 \\ 
& SIVI \cite{yin2018semi} & 0.7619 & 0.7846 & 0.7662
 \\ 
&  ADVI \cite{kucukelbir2017automatic} & 0.7763 & 0.6846 & 0.7492 \\
&  NUTS (MCMC) \cite{hoffman2014no} & 0.7907 & 0.7183 & 0.7549\\
\hline
 \multirow{6}{*}{2021 Haiti earthquake} & \textbf{SVGDM} & \textbf{0.9650} & \textbf{NGA} & \textbf{0.9687}
\\
& VCBI \cite{xu2022seismic} & 0.9123 & NGA & 0.9123\\
 & DisasterNet \cite{li2023disasternet} & 0.9421 & NGA & 0.9410\\
   & Prior Model \cite{zhu2015geospatial, nowicki2018global}  & 0.8712 & NGA & - \\
&  BBVI \cite{ranganath2014black} & 0.7729 & NGA & 0.8125
 \\ 
& SIVI \cite{yin2018semi} & 0.7964 & NGA & 0.8123
 \\ 
&  ADVI \cite{kucukelbir2017automatic} & 0.8155 & NGA & 0.8222 \\
&  NUTS (MCMC) \cite{hoffman2014no} & 0.7612 & NGA & 0.7747\\
\hline
 \multirow{6}{*}{2023 Turkey-Syria earthquake} & \textbf{SVGDM} & \textbf{NGA} & \textbf{NGA} & \textbf{0.9888}
\\
 & VCBI \cite{xu2022seismic} & NGA & NGA & 0.9361
 \\ 
 & DisasterNet \cite{li2023disasternet} & NGA & NGA & 0.9564\\
  & Prior Model \cite{zhu2015geospatial, nowicki2018global}  & NGA & NGA & 0.9051\\
 & AdaBoost\cite{patten2024data} & - & - & 0.9300 \\
&  BBVI \cite{ranganath2014black} & NGA & NGA & 0.8233
 \\ 
& SIVI \cite{yin2018semi} & NGA & NGA & 0.7947
 \\ 
&  ADVI \cite{kucukelbir2017automatic} & NGA & NGA & 0.7315 \\
&  NUTS (MCMC) \cite{hoffman2014no} & NGA & NGA & 0.8013\\
 \toprule
\end{tabular}
}
\label{AUC_real}
\end{center}
\end{table*}

\begin{figure*}[ht]
  \centering
\includegraphics[width=0.8\linewidth]{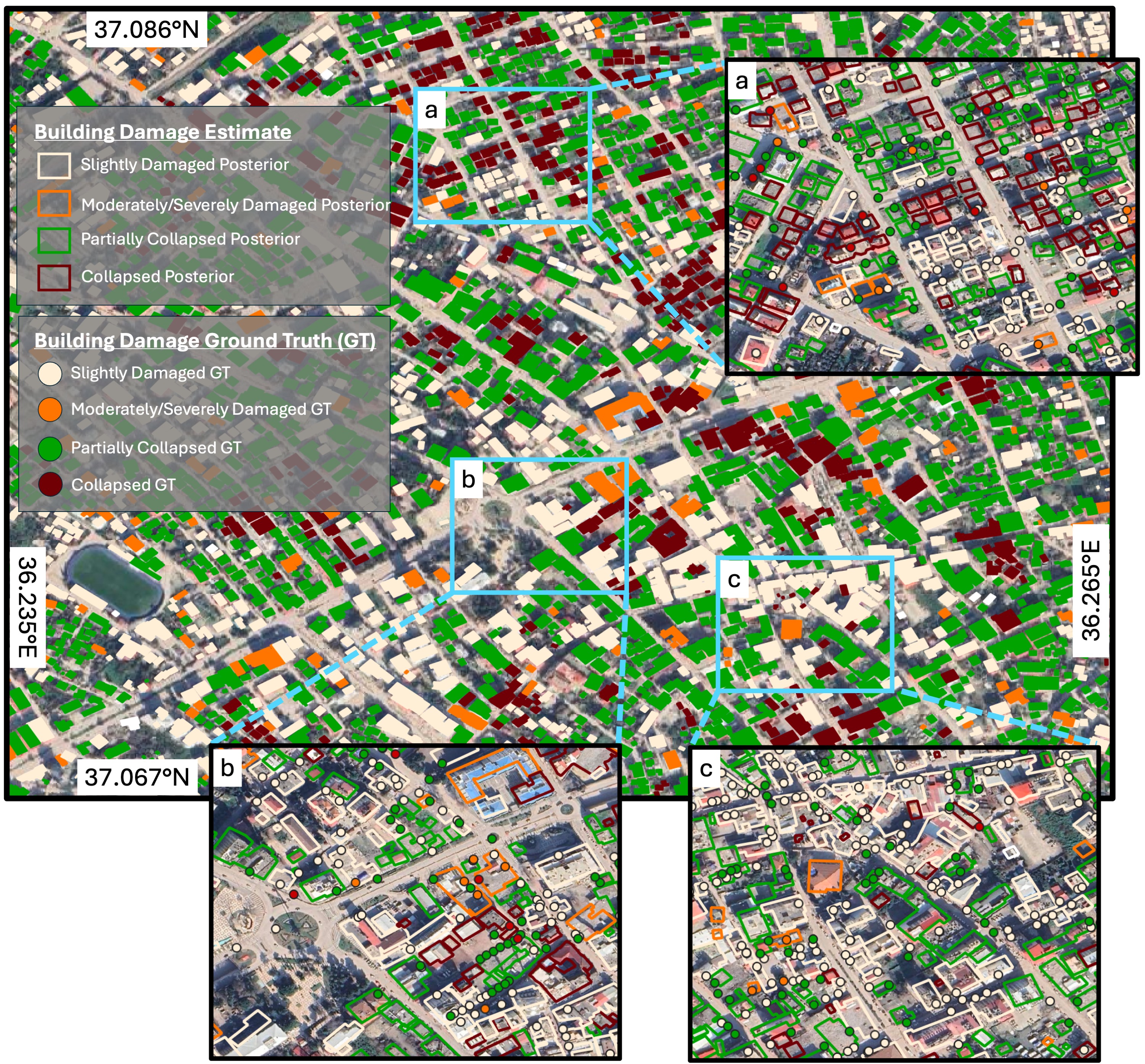}
  \caption{\textbf{Estimated building damage probability maps in Osmaniye, Turkey}. Figures (a)–(c) present the building damage estimates with ground truth information. The legend colors represent the building damage levels, as indicated.}
  \label{EQ_BDs}
    \end{figure*}
    
\subsection{Evaluation of Seismic Multi-Hazard Estimation over Multiple Seismic Events}

We evaluate our framework on three major earthquake events: the 2020 Puerto Rico earthquake, the 2021 Haiti earthquake, and the 2023 Turkey-Syria earthquake. These events represent diverse scenarios with different geological settings, building typologies, and data availability conditions, providing a comprehensive test of capabilities of our framework. Unlike traditional approaches that model each hazard independently, our framework enables simultaneous estimation of multiple earthquake-induced hazards and their cascading effects. As sudden-onset events, earthquakes cause damage almost instantaneously, making this a static rather than temporal prediction task.

Table \ref{AUC_real} presents the area under the curve (AUC) scores for three latent variables: landslide ($z_{LS}$), liquefaction ($z_{LF}$), and building damage ($z_{BD}$). The joint estimation of these interconnected hazards allows our model to capture the complex interactions between ground shaking, soil conditions, and structural vulnerability. Our framework consistently outperforms existing methods across all three earthquake events, demonstrating its robustness across different scenarios and data conditions. For the 2020 Puerto Rico earthquake, SVGDM achieves AUC scores of 0.9331, 0.9317, and 0.9512 for landslide, liquefaction, and building damage estimation respectively. These consistently high scores across all three hazards demonstrate the framework's ability to maintain accuracy while simultaneously modeling multiple interrelated phenomena. These results show significant improvements over both traditional probabilistic methods (BBVI, SIVI, ADVI, NUTS) and recent deep statistical learning approaches (VCBI, DisasterNet). The performance gap is particularly notable compared to classical variational inference methods, where our approach shows a 15-20\% improvement in AUC scores. For instance, compared to BBVI, our method achieves improvements of 14.19\%, 15.86\%, and 20.89\% for landslide, liquefaction, and building damage respectively.

The 2021 Haiti earthquake results further demonstrate the robustness of our framework, with AUC scores of 0.9550 for landslide susceptibility and 0.9587 for building damage estimation. While liquefaction ground truth was not available (NGA) for this event due to limited field surveys, the strong performance in other aspects suggests the ability of the framework to handle partially observed scenarios. The consistent improvement over baseline methods (ranging from 3-18\% higher AUC) highlights the advantage of our multi-resolution approach. Notably, our framework maintains high performance even with the limited availability of high-resolution background knowledge in the Haiti context, showing its adaptability to different data conditions.

For the 2023 Turkey-Syria earthquake, while ground truth for landslide and liquefaction was not available (NGA), our framework achieves an AUC score of 0.9488 for building damage estimation. This performance surpasses both traditional methods like BBVI (0.8233, improved by 12.55\%) and more recent approaches like VCBI (0.9025, improved by 4.63\%) and DisasterNet (0.9315, improved by 1.73\%). Our method also outperforms the Prior Model (0.9391) and AdaBoost (0.9300), which are designed for building damage assessment and incorporate domain-specific engineering knowledge. This robust performance, achieved with partial ground truth availability, demonstrates the effectiveness of our framework in real-world scenarios where complete validation data may not be available.

Figure \ref{EQ_BDs} provides a visual demonstration of our building damage estimation capabilities through the 2023 Turkey-Syria earthquake case study in Osmaniye. To enhance the interpretation of building damage probability maps \cite{li2024rapid}, we categorize the building damage probability into four levels: slight damage (including no damage) (probability $\leq$ 0.3), moderate damage (0.3 $<$ probability $\leq$ 0.65), partial collapse (0.65 $<$ probability $\leq$ 0.8), and collapse (probability $>$ 0.8). We adjust the threshold to align optimally with the limited available ground truth information. The results show that our estimated damage patterns closely align with ground truth observations across different damage levels, from slight damage (white) to complete collapse (dark red). Our framework successfully captures the spatial distribution of damage severity, particularly evident in areas with mixed damage patterns where buildings in close proximity experienced different levels of damage. The high-resolution predictions (shown in insets a-c) demonstrate the ability of the model to distinguish between adjacent buildings with different damage states, highlighting the effectiveness of our multi-resolution approach in preserving fine-grained spatial details while maintaining global consistency in the damage assessment.

The comprehensive nature of our framework's hazard estimation capabilities is particularly evident in the Puerto Rico earthquake case study. Our framework's robust performance extends to the 2021 Haiti earthquake, as shown in Supplementary Figure \ref{earthquake_diff_setting}. The visualization presents building damage estimates across a densely populated urban area with complex architectural patterns. The model accurately captures the heterogeneous damage distribution, with particularly strong performance in identifying clusters of moderate to severe damage alongside collapsed structures. The high-resolution predictions, detailed in insets Figure \ref{earthquake_diff_setting} (a)-(c), demonstrate the ability of our model to differentiate damage states between adjacent buildings despite the challenging urban layout typical of built environment in Haiti. This precise discrimination of damage levels is especially evident in mixed-damage zones where neighboring buildings experienced markedly different impacts from the earthquake. The close correspondence between predicted damage patterns and ground truth observations validates our quantitative results (AUC: 0.9550) and highlights the effectiveness of our model in scenarios with diverse building typologies and construction practices.

In addition, we also show our results for the reconstruction of buildings for the Puerto Rico earthquake in the Supplementary Figure \ref{PR_BD}, where the model effectively identifies the varying levels of damage in different urban contexts. This analysis, combined with our liquefaction predictions, showcases the model's ability to simultaneously assess multiple aspects of seismic risk. The predictions demonstrate strong agreement with ground truth observations across all damage categories, from no/slightly damaged structures (orange) to collapsed buildings (dark red). The model successfully captures both isolated damage instances and larger damage clusters, reflecting the complex spatial patterns of seismic impacts.

Supplementary Figure \ref{PR_LF} further demonstrates the effectiveness of our model in liquefaction prediction for the Puerto Rico earthquake. The simultaneous assessment of liquefaction susceptibility alongside building damage enables a more comprehensive understanding of seismic impacts. The visualization shows estimated liquefaction probability maps with ground truth labels across different geological settings, from coastal areas to inland regions. The close alignment between predicted liquefaction zones (shown in blue intensity) and ground truth observations (pink circle) validates our model's ability to capture the spatial distribution of soil liquefaction susceptibility. This is particularly evident in areas where geological conditions and ground shaking characteristics combine to create heightened liquefaction risk. The consistent performance across both structural and geotechnical hazards demonstrates the versatility of our joint estimation approach.

\subsection{Evaluation of Hurricane-Induced Damage Estimation for the 2023 Hurricane Ian}

In this section, we evaluate our framework on the 2023 Hurricane Ian case. While hurricanes are inherently temporal events with evolving dynamics over time, our analysis focuses on a static assessment of building damage due to data availability constraints of post-disaster data. Satellite-based Damage Proxy Maps (DPMs), which serve as our primary source of building damage information, are typically collected only after the hurricane has passed, providing a single snapshot of the final damage state. This limitation in temporal data availability necessitates a static modeling approach, though we acknowledge that incorporating time-series data could potentially enhance our understanding of damage progression during the event. Despite this constraint, in this section, we demonstrate that our framework achieves robust performance in assessing hurricane-induced building damage.

As shown in Table \ref{tab:model_comparison}, SVGDM demonstrates superior accuracy in assessing hurricane-induced building damage in Lee County, Florida, achieving balanced detection capabilities with a True Positive Rate of 0.8217 and True Negative Rate of 0.8197. This balanced performance translates to an AUC score of 0.8123, indicating the robust ability of our model to accurately distinguish between damaged and undamaged buildings. The results demonstrate that our model successfully identifies over 82\% of damaged buildings while maintaining a similar level of accuracy in correctly classifying undamaged structures, demonstrating its reliability across different building conditions.

\begin{table}[htbp]
    \centering
    \caption{Performance comparison of building damage assessment models for hurricane-induced damage detection for hurricane Ian case study in Lee County, Florida.}
    \begin{tabular}{lccc}
        \toprule
        \textbf{Model} & \textbf{TPR} & \textbf{TNR} & \textbf{AUC} \\
        \midrule
        SVGDM$^*$ & \textbf{0.8217} & \textbf{0.8197} & \textbf{0.8123} \\
        Bayesian network-based Model$^*$ & 0.8293 & 0.6221 & 0.7553 \\
        FCS-Net (w/o finetuning) & 0.2713 & 0.8941 & -- \\
        FCS-Net (w/ finetuning) & 0.2098 & 0.9386 & -- \\
        Dual-HRNet (w/o finetuning) & 0.0912 & 0.9795 & -- \\
        Dual-HRNet (w/ finetuning) & 0.8217 & 0.6251 & -- \\
        DPM-based Model$^*$ & 0.6498 & 0.6249 & 0.6739 \\
        Fragility Curve$^*$ & 0.5669 & 0.6246 & 0.5695 \\
        \bottomrule
        \multicolumn{4}{l}{\footnotesize $^*$Models trained without labels} \\
        \multicolumn{4}{l}{\footnotesize TPR: True Positive Rate, TNR: True Negative Rate, AUC: Area Under Curve} \\
        \multicolumn{4}{l}{\footnotesize DPM: Damage Proxy Map, w/: with, w/o: without} \\
        \multicolumn{4}{l}{\footnotesize Finetuning: Model adaptation using Hurricane Ian labeled data} \\
        \multicolumn{4}{l}{\footnotesize '--' indicates metric not available for deterministic models}
    \end{tabular}
\label{tab:model_comparison}
\end{table}

When compared to existing approaches documented in \cite{wang2024scalable}, the performance of Temporal-SVGDM is noteworthy across different categories of methods. Traditional methods like the Fragility Curve model (TPR: 0.5669, TNR: 0.6246) and DPM-based Model (TPR: 0.6498, TNR: 0.6249) show lower and imbalanced performance. Deep learning approaches exhibit interesting patterns: without finetuning, both FCS-Net and Dual-HRNet show extremely low TPR (0.2713 and 0.0912 respectively) despite high TNR, indicating a strong bias toward predicting buildings as undamaged. Even with finetuning using Hurricane Ian labeled data, the performance of FCS-Net remains imbalanced (TPR: 0.2098, TNR: 0.9386), while Dual-HRNet achieves comparable TPR (0.8217) but at the cost of reduced TNR (0.6251). The Bayesian network-based Model, while achieving similar TPR (0.8293), shows lower TNR (0.6221), resulting in a lower AUC of 0.7553.

A key advantage of Temporal-SVGDM is its ability to achieve this high performance without requiring labeled training data, as indicated by the asterisk in Table \ref{tab:model_comparison}. This is particularly significant for rapid disaster response scenarios where obtaining ground truth labels is time-consuming and often impractical. The balanced performance of Temporal-SVGDM across both TPR and TNR, combined with its label-free learning capability, makes it particularly suitable for real-world deployment in hurricane damage assessment.

\begin{figure*}[ht]
  \centering
\includegraphics[width=1.1\linewidth]{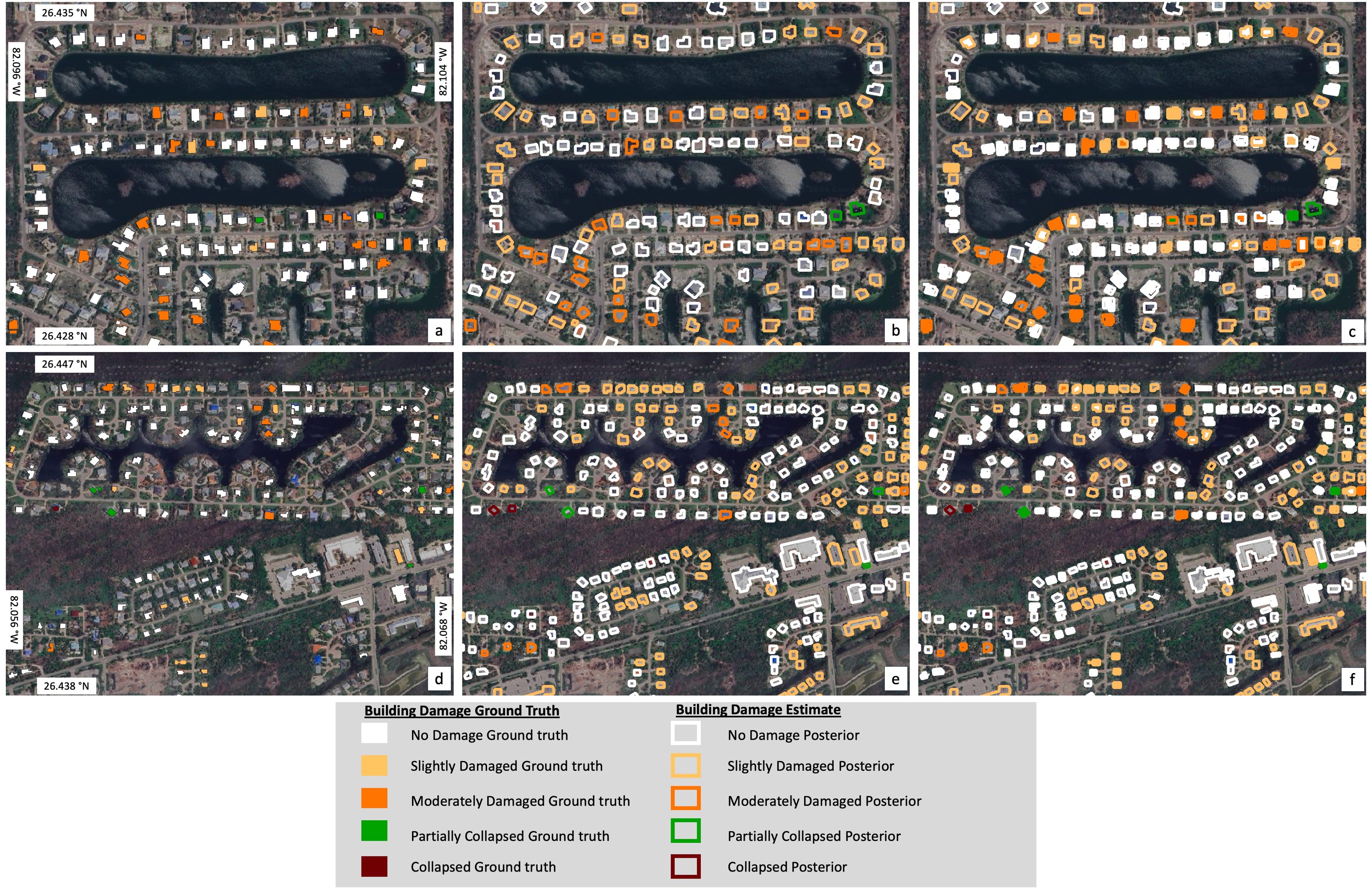}
  \caption{\textbf{Estimated building damage probability maps for Hurricane Ian case study in Lee County, Florida.} Figures (a)–(c) present the building damage ground truth, building damage estimation, and their overlay for the region from 82.096°W to 82.104°W and 26.238°N to 26.435°N. Figures (d)–(f) present the corresponding views for the region from 82.056°W to 82.068°W and 26.438°N to 26.447°N. Colors in the legend indicate building damage levels.}
  \label{Hurricane_BD}
    \end{figure*}

The effectiveness of our model is visually demonstrated in Figure \ref{Hurricane_BD}, which presents building damage assessment results for two distinct regions in Lee County, Florida. To enhance the interpretation of building damage probability maps \cite{li2024rapid}, we categorize the building damage probability into five levels: no damage ((probability $\leq$ 0.15), slight damage (0.15 $<$ probability $\leq$ 0.5), moderate damage (0.5 $<$ probability $\leq$ 0.7), partial collapse (0.7 $<$ probability $\leq$ 0.85), and collapse (probability $>$ 0.85). The comparison between ground truth (Figure \ref{Hurricane_BD} (a)(d)) and our estimation (Figure \ref{Hurricane_BD} (b)(e)) shows consistency in damage level assessment across different areas. The overlay visualization (Figure \ref{Hurricane_BD} (c)(f)) particularly highlights the ability of our method to capture varying degrees of damage, from no damage (shown in white outlines) to collapse (shown in dark red outlines). This visual validation complements our quantitative metrics and confirms the practical utility of our model for post-hurricane damage assessment.

\subsection{Evaluation of Wild Fire Spread Estimation}

We also evaluate the performance of Temporal-SVGDM for wildfire spread prediction through comprehensive comparison with several baseline approaches, including both traditional statistical methods and deep learning models documented in \cite{gerard2023wildfirespreadts}. As shown in Table \ref{tab:results}, the results demonstrate the significant advantages of our framework across multiple performance metrics. Temporal-SVGDM achieves superior performance with an F1 score of 0.5913 and Average Precision (AP) of 0.4430, significantly outperforming all baseline approaches. This performance can be attributed to the ability of our framework to capture both the spatial dependencies and temporal evolution patterns inherent in wildfire spread dynamics. The substantial margin of improvement over traditional methods is particularly noteworthy. For example, our F1 score exceeds that of logistic regression (0.432) by approximately 37\%, indicating the enhanced capability of Temporal-SVGDM in handling the complex, non-linear nature of fire spread patterns.

When compared to deep learning approaches, Temporal-SVGDM maintains its performance advantage. Both mono-temporal and multi-temporal ConvLSTM models achieve F1 scores of 0.310, demonstrating the limitations of conventional sequence modeling approaches in capturing the intricate dynamics of wildfire spread. Even more sophisticated architectures like U-Net (F1: 0.341) and UTAE (F1: 0.350) fall significantly short of the performance of temporal-SVGDM, highlighting the benefits of our causal modeling approach in understanding and predicting wildfire behavior. The consistently higher performance across both metrics underscores the robust ability of Temporal-SVGDM to model the complex relationships between various environmental factors and fire progression. This is particularly significant given the challenging nature of wildfire spread prediction, where multiple interacting factors must be considered simultaneously across different spatial and temporal scales.

\begin{table}[ht]
\centering
\caption{\textbf{Comparison of model performance on wildfire spread prediction.} AP stands for Average Precision.}
\begin{tabular}{lcc}
\hline
\textbf{Model} & \textbf{F1 Score} & \textbf{AP} \\
\hline
SVGDM & \textbf{0.5913} & 
\textbf{0.4430} \\
Logistic Regression & 0.432 & 0.279 \\
U-Net & 0.341 & 0.341 \\
ConvLSTM (Mono-temporal) & 0.310 & 0.292 \\
ConvLSTM (Multi-temporal)  & 0.310 & 0.306 \\
UTAE & 0.350 & 0.372 \\
\hline
\end{tabular}
\label{tab:results}
\end{table}

Supplementary Figure \ref{ActiveFire} provides a visual demonstration of our model's predictive capabilities through a side-by-side comparison of ground truth labels, our model's predictions, and benchmark model predictions \cite{gerard2023wildfirespreadts} across six consecutive time steps. Our model demonstrates strong advantages in both spatial accuracy and temporal consistency. The predictions show particularly good performance in capturing the size and shape of fire clusters, as evident in steps a2 and a3 where our model accurately identifies both the presence and extent of major fire events. In contrast, the benchmark predictions tend to underestimate the fire extent and miss some significant fire clusters.

The spatial precision of our predictions is especially notable in capturing scattered fire patterns. For example, in time step a3, our model successfully identifies both the upper and lower fire clusters, maintaining their relative sizes and positions, while the benchmark model only partially captures these patterns. The model also shows robust performance in tracking fire evolution over time, successfully identifying the appearance of new fire locations (as seen in a5 and a6) and the disappearance of extinguished fires (as demonstrated in a4). 

These visualization results complement our quantitative metrics, providing clear visual evidence for why Temporal-SVGDM achieves superior F1 and AP scores compared to baseline approaches. The model's ability to accurately capture both the spatial distribution and temporal evolution of fires, while maintaining lower false positive rates than the benchmark model, demonstrates the effectiveness of our multi-resolution, causality-aware approach in handling the complex, dynamic nature of wildfire spread.

\subsection{Analysis of Model Performance Under Different Background Knowledge Conditions}

\begin{table*}[htbp]
\begin{center}
\caption{\textbf{This table presents the experimental scenarios for evaluating Temporal-SVGDM under different background knowledge conditions in earthquake cases.} NGA means no ground truth label is available. Background knowledge configurations: VF-BK (Varying-resolution Full Background Knowledge); LF-BK (Low-resolution Full Background Knowledge); LP-BK (Low-resolution Partial Background Knowledge)}
\scalebox{1}{
\begin{tabular}{c|c|ccc}
\toprule
\textbf{Earthquake} & \textbf{Description} & \textbf{$z_{\text{LS}}$} & \textbf{$z_{\text{LF}}$} & \textbf{$z_{\text{BD}}$}\\
\midrule
\multirow{3}{*}{2020 Puerto Rico earthquake} & \textbf{VF-BK} & \textbf{0.9331} & \textbf{0.9317} & \textbf{0.9512} \\
& LF-BK & 0.9214 & 0.9201 & 0.9297 \\
& LP-BK & 0.8918 & 0.8893 & 0.9004 \\
\midrule
\multirow{3}{*}{2021 Haiti earthquake} & \textbf{VF-BK} & \textbf{0.9550} & \textbf{NGA} & \textbf{0.9587} \\
& LF-BK & 0.9307 & NGA & 0.9249 \\
& LP-BK & 0.8925 & NGA & 0.8993 \\
\midrule
\multirow{3}{*}{2023 Turkey-Syria earthquake} & \textbf{VF-BK} & \textbf{NGA} & \textbf{NGA} & \textbf{0.9488} \\
& LF-BK & NGA & NGA & 0.9233 \\
& LP-BK & NGA & NGA & 0.8832 \\
\bottomrule
\end{tabular}
}
\label{earthquake_diff_setting}
\end{center}
\end{table*}

To systematically evaluate how varying levels of data availability and resolution affect model performance, we conducted comprehensive experiments across three types of natural disasters - earthquakes, hurricanes, and wildfires. Each disaster type presents unique challenges in terms of temporal dynamics, causal relationships, and data availability patterns. We evaluated Temporal-SVGDM under three background knowledge scenarios: varying-resolution full background knowledge (VF-BK) utilizing all available resolutions, low-resolution full background knowledge (LF-BK) using only lower resolution data sources, and low-resolution partial background knowledge (LP-BK) using limited data. This design allowed us to test the model's performance even when background knowledge was constrained.

Our model demonstrates robust adaptability across different data availability conditions in earthquake scenarios. For the VF-BK scenario, we incorporate multi-resolution background knowledge (Landslide: 250m and 1000m; Liquefaction: 500m and 1000m; Building Damage: 30m), where Temporal-SVGDM shows superior performance. In the 2020 Puerto Rico earthquake case, as shown in Table \ref{earthquake_diff_setting}, the model achieved AUC scores of 0.9331, 0.9317, and 0.9512 for landslide, liquefaction, and building damage estimation respectively, outperforming recent deep learning approaches like VCBI (0.9012, 0.9034, 0.9123) and DisasterNet (0.9293, 0.9284, 0.9413). When transitioning to the LF-BK scenario using only lower resolution data, performance remains robust with only a 1-2\% reduction in AUC scores, still maintaining substantial advantages over traditional methods like BBVI (0.7912, 0.7731, 0.7423) and SIVI (0.7619, 0.7846, 0.7662). Even in the most challenging LP-BK scenario with only partial background knowledge available, the model maintains AUC scores consistently above 0.88 - notably higher than sophisticated baselines like NUTS (MCMC) which achieves scores around 0.75-0.79.

We also test the performance of our model in the hurricane case. As shown in Figure \ref{BNs}(b), each hurricane-related variable has a single native resolution of background knowledge. In the VF-BK scenario utilizing all available information, our model achieves balanced detection capabilities with a TPR of 0.8217, TNR of 0.8197, and AUC of 0.8123, substantially outperforming traditional approaches like the Fragility Curve model (TPR: 0.5669, TNR: 0.6246) and deep learning methods like FCS-Net even with finetuning (TPR: 0.2098, TNR: 0.9386). The transition to LP-BK shows notable but manageable performance degradation (TPR: 0.7073, TNR: 0.6525, AUC: 0.6891), yet still surpasses DPM-based Models (TPR: 0.6498, TNR: 0.6249, AUC: 0.6739) and remains competitive with more complex approaches requiring labeled training data.

We further evaluate our model on wildfire spread prediction case, which features intricate temporal dependencies and multiple interacting factors. Under the VF-BK scenario, the model achieves an F1 score of 0.5913 and Average Precision of 0.4430, significantly outperforming both traditional methods like Logistic Regression (F1: 0.432, AP: 0.279) and sophisticated deep learning approaches like U-Net (F1: 0.341, AP: 0.341) and ConvLSTM variants (F1: 0.310). Even in the LP-BK scenario with reduced background knowledge, our model maintains F1 and AP scores of 0.4613 and 0.3710 respectively, still surpassing all baseline methods, including UTAE (F1: 0.350, AP: 0.372). This robust performance across different background knowledge conditions demonstrates the ability of our model to effectively leverage available information while maintaining resilience to data limitations.

Comparing performance degradation patterns across disaster types reveals interesting insights about the adaptability of our model. In earthquake scenarios, the drop from VF-BK to LP-BK is relatively moderate (approximately 4-5\% in AUC), while hurricane assessment shows a more substantial decline (12.32\% in AUC). This difference likely stems from the complementary nature of multiple resolution sources in earthquake assessment versus the distinct roles of flood and wind information in hurricane damage prediction. The wildfire case shows an intermediate degradation pattern (13\% in F1 score but only 7.2\% in AP), suggesting that temporal dependencies help maintain prediction rankings even with reduced spatial information.

These varying degradation patterns across disaster types have important implications for practical deployment. For earthquake assessment, the resilience of our model to resolution reduction suggests it can be effectively deployed even in regions with limited high-resolution data availability. In hurricane scenarios, the results emphasize the importance of maintaining comprehensive hazard information for optimal performance, though the model remains functional with limited data. For wildfire prediction, the maintained ranking ability under limited information suggests the model could be valuable for prioritizing response efforts even when full sensor coverage is unavailable. The consistent outperformance of baseline methods across all background knowledge conditions suggests that the core architectural choices of our model and provide fundamental advantages independent of data availability. This is evidenced by the ability of our model to maintain balanced performance metrics (TPR/TNR in hurricane cases, precision/recall in wildfire prediction) even under reduced information conditions, whereas baseline methods often show severe imbalances in their predictions when faced with limited data.


\section{Discussion}
This work introduces Temporal-SVGDM, a novel framework that demonstrates remarkable versatility in modeling complex systems with multi-resolution background knowledge and causal dependencies. Our comprehensive evaluation across three distinct disaster types - earthquakes, hurricanes, and wildfires - not only validates the effectiveness of Temporal-SVGDM but also reveals fundamental insights about modeling systems with different temporal characteristics.

A fundamental contribution of our work is the framework's ability to effectively integrate and process multi-resolution background knowledge while preserving causal relationships. Unlike traditional approaches that force uniform resolution through interpolation or downsampling, Temporal-SVGDM allows each data source to maintain its native resolution through resolution-specific SDEs coupled with a causal score mechanism. This approach provides several key advantages: 1) It preserves fine-grained details from high-resolution sources while simultaneously capturing broader patterns from lower-resolution data, 2) It enables the framework to handle missing or sparse data at certain resolutions by leveraging information from other scales, and 3) It maintains the physical meaning of causal relationships across different resolutions through carefully designed score matching and diffusion processes.
The effectiveness of this multi-resolution approach is particularly evident in our earthquake case studies, where the model successfully integrates landslide susceptibility data at both 250m and 1000m resolutions with building damage assessments at 30m resolution. The framework's ability to maintain performance even when high-resolution data is limited (as shown in the LP-BK scenario results) demonstrates the robustness of our multi-resolution design. This capability represents a significant advance over existing methods that either require uniform resolution or struggle to maintain causal consistency across different scales.

Our results demonstrate that explicitly modeling causal relationships while preserving resolution-specific information yields substantial advantages over traditional approaches that either ignore causality or force uniform resolution. The performance of our model across all three disaster types (earthquake AUC: above 0.95, hurricane AUC: 0.81, wildfire F1: 0.59) suggests that maintaining the native resolution of different data sources, rather than forcing interpolation to a common scale, better captures the underlying physical processes. This finding challenges the common practice of resolution standardization in environmental modeling and suggests that preserving multi-resolution relationships may be crucial for accurate system representation.
The performance of our model under varying background knowledge conditions provides insights into the robustness-resolution trade-off in complex system modeling. The observed degradation patterns - from modest in earthquakes (4-5\% AUC reduction) to more pronounced in hurricanes (12.32\%) and wildfires (13\%) - reveal how different types of physical processes respond to information reduction. This varying sensitivity likely reflects the underlying physical characteristics of each system: earthquake damage patterns exhibit stronger spatial correlation that helps maintain performance even with reduced data, while the dynamic nature of wildfires makes them more sensitive to missing temporal information.

The ability of our framework to maintain balanced performance metrics (TPR/TNR in hurricane assessment, precision/recall in wildfire prediction) even under reduced information conditions has important implications for operational deployment. This characteristic suggests that the model could provide reliable predictions even in data-sparse regions or during sensor network failures, a crucial capability for real-world disaster monitoring systems. However, several important challenges remain. While our framework handles varying spatial resolutions effectively, temporal resolution differences still pose challenges, particularly in cases with dramatically different sampling frequencies. Future work could explore adaptive temporal diffusion mechanisms that better handle such scenarios. Additionally, the computational complexity of processing multi-resolution data suggests the need for more efficient algorithms for real-time applications.

Another key contribution of our work is the ability of our model to handle both static and temporal scenarios within a unified approach. In earthquake modeling, where causal chains operate almost instantaneously, Temporal-SVGDM effectively captures the immediate cascading effects of ground shaking on landslides, liquefaction, and building damage. The high performance in this static scenario (AUC: Above 0.95) demonstrates the capability of our model to capture complex spatial dependencies and instantaneous causal relationships. Conversely, in wildfire prediction, where the system evolves over time with changing environmental conditions, our model successfully captures the temporal progression of fire spread while maintaining causal relationships between fuel conditions, weather patterns, and topography (F1: 0.59). This dual capability - handling both immediate causal effects and temporal evolution - represents a significant advance over existing methods that typically specialize in either static or temporal modeling.

Looking forward, the principles demonstrated in this work could extend beyond disaster modeling to other domains with similar challenges in multi-resolution data integration. Climate science, environmental monitoring, and urban systems analysis all face comparable challenges in synthesizing heterogeneous data sources while maintaining causal relationships. The success of our score-based diffusion approach in handling these challenges suggests promising directions for these broader applications. More broadly, our results indicate that the future of complex system modeling may lie not in forcing standardization of heterogeneous data, but in developing frameworks that can naturally handle multi-resolution information while preserving causal relationships. This perspective could influence how we approach data collection and sensor network design for environmental monitoring, suggesting that maintaining multiple resolution streams may be more valuable than pursuing uniform high-resolution coverage.

\section{Methods}

In this section, we detail the key components of Temporal Score-based Variational Graphical Diffusion for Multi-resolution background knowledge (Temporal-SVGDM). We formulate complex causal systems with multi-resolution background knowledge as a new family of BNs. We implement multiple coupled SDEs corresponding to nodes in these BNs, allowing us to maintain native resolutions of diverse data sources, perform multi-SDE diffusion across the entire causal graph, integrate multi-resolution data through forward diffusion processes for each hidden variable, and utilize conditional causal scores to incorporate information from parent nodes. The final goal of this work is to estimate the posteriors of latent variables and infer quantitative causal dependencies, effectively balancing fine-grained details with overall patterns while maintaining causal integrity throughout the modeling process.

\subsection{Score-based Generative Model through SDEs}

Score-based generative models, which is a type of continuous diffusion model, have emerged as powerful tools for learning complex, high-dimensional data distributions. These models leverage the score function, which is the gradient of the log-density of the data, to generate new samples by simulating a reverse diffusion process. The foundational idea was introduced by Song and Ermon in their seminal works on generative modeling using score matching and diffusion processes~\cite{song2019generative, song2020score}. In a score-based generative model, the data distribution is gradually perturbed by adding noise through a forward diffusion process, resulting in a sequence of increasingly noisy data representations. The reverse diffusion process then reconstructs the original data distribution by denoising these representations, guided by the learned score function. Mathematically, the forward and reverse processes are typically described using SDEs, making it a type of continuous diffusion model.

\noindent\textbf{Forward Diffusion Process:} The forward diffusion process is:

\begin{equation}
dz = f(z, t) dt + g(t) dW
\end{equation}

\noindent where $z$ is the data, $f(z, t)$ is the drift term, $g(t)$ is the diffusion term, and $W$ is the Wiener process (standard Brownian motion).

\noindent\textbf{Reverse Diffusion Process:} The reverse diffusion process is described by the reverse-time SDE:

\begin{equation}
dz = [f(z, t) - g(t)^2 \nabla_z \log p_t(z)] dt + g(t) d\bar{W}
\end{equation}

where $\nabla_z \log p_t(z)$ is the score function, representing the gradient of the log-density of the data at time $t$, and $\bar{W}$ is the reverse-time Wiener process.
Training a score-based diffusion model involves learning the score function $\nabla_z \log p_t(z)$ using score matching techniques. The learned score function is then used to guide the reverse diffusion process for generating new data samples.

\subsection{Variational Inference}
\label{VIIntro}
Variational inference (VI) is a widely-used technique for approximating complex posterior distributions in Bayesian models. It transforms the inference problem into an optimization task by introducing a variational distribution $q_{\phi}(z|y)$ to approximate the true posterior $p(z|y)$. The goal is to find the parameters $\phi$ that make $q_{\phi}(z|y)$ as close as possible to $p(z|y)$. The quality of the approximation is measured by the ELBO, which serves as a proxy for the marginal likelihood. Maximizing the ELBO is equivalent to minimizing the Kullback-Leibler (KL) divergence between the variational distribution and the true posterior. The ELBO is defined as: 

\begin{equation*}
\mathcal{L} = \mathbb{E}_{z\sim q_{\varphi}(z|y)}[\log p(y|z) + \log p(z) - \log q_{\varphi}(z|y)]
\end{equation*}

\noindent where $p(y|z)$ is the likelihood, $p(z)$ is the prior, and $q_{\varphi}(z|y)$ is the variational distribution.

Besides, the optimization objective for variational inference is to maximize the ELBO with respect to the parameters $\varphi$: $\varphi^* = \arg \max_{\varphi} \text{ELBO}$. This optimization is typically performed using gradient-based methods, with gradients of the ELBO computed using techniques such as the reparameterization trick or score function estimators.

\subsection{Multi-resolution Temporal Causal Graphical Diffusion Modeling} 
\subsubsection{Multi-Resolution Temporal Inference and Baseline}\label{Baseline}

Traditional approaches in multi-resolution scenarios often overlook resolution-specific nuances by directly stacking background knowledge \cite{de2011multi,xu2014assimilation,atkinson2013downscaling}. Existing multi-resolution methods like super-resolution networks \cite{saharia2022image} and multi-scale fusion techniques \cite{song2019survey} focus primarily on spatial relationships without considering temporal evolution. Meanwhile, temporal modeling approaches such as recurrent neural networks \cite{reichstein2019deep}, temporal convolutional networks \cite{torres2021deep}, and transformer-based architectures \cite{vaswani2017attention} typically assume consistent observational resolution across time steps. Recent attempts at spatio-temporal modeling \cite{shi2015convolutional,yu2017spatio} have shown promise but struggle with varying resolutions. Earth system modeling approaches \cite{raissi2019physics,brunton2016discovering} have explored multi-scale phenomena but generally require uniform temporal sampling.

To address these multi-resolution challenges, the field of probabilistic modeling and inference has developed several methods to handle complex distributions and uncertainty quantification. Classical approaches include Black Box Variational Inference (BBVI) \cite{ranganath2014black}, Semi-Implicit Variational Inference (SIVI) \cite{yin2018semi}, Automatic Differentiation Variational Inference (ADVI) \cite{kucukelbir2017automatic}, and No-U-Turn Sampler (NUTS) \cite{hoffman2014no}. BBVI uses Monte Carlo gradient estimates, SIVI employs implicit distributions for expressive posteriors, and ADVI automates VI through automatic differentiation. NUTS, while not a VI method, serves as a strong baseline with its adaptive sampling. While recent work in score-based diffusion models \cite{song2020improved,song2021maximum,dhariwal2021diffusion} and variational inference \cite{kingma2021variational,yang2023diffusion} has advanced the field, existing methods struggle to fully capture multi-resolution dynamics, temporal evolution, and causal relationships in complex systems. Particularly, current approaches either focus on temporal consistency \cite{shi2015convolutional,yu2017spatio} or spatial resolution \cite{reichstein2019deep,raissi2019physics}, but rarely both. Our proposed framework addresses these limitations by integrating SDEs, score-based diffusion models, and variational inference, enabling comprehensive probabilistic inference over temporal multi-resolution systems with causal dependencies.

\noindent\textbf{Causal Bayesian Network.} We consider our target area as a continuous space $L$, which can be discretized at different resolutions. Given a sequence of temporal background knowledge $Y^s$ where $s \in {1,\cdots, S}$ is the sequence length ($S=1$ for non-temporal case), for each unobserved variable $z_i$, we denote $z_i^{l,s} \in Z$ as the value of the $i^{th}$ hidden variable at spatial location $l \in \mathcal{L}$ and temporal step $s$. The corresponding background knowledge are denoted as $y_i^{k_i, l', s} \in Y$, where $k_i$ represents the resolution of the background knowledge, and $l'$ is the location in the discretized space at resolution $k_i$.

In the complex system, $y_{i}^{k_i,s}$ represents one background information of the $i^{\text{th}}$ variable at resolution $k_i$ and time step $s$. For example, this could be satellite imagery for ground deformation at a specific spatial resolution, or ground failure model outputs for landslides and liquefaction generated at a particular grid size in earthquake case. These multi-resolution background knowledge, typically coming from different sources, allow us to capture different aspects of the complex system at varying scales and time steps, representing both fine-grained details and broader patterns in both space and time. $K_i$ denotes the number of different resolutions at which background knowledge in finest resolution is available for $z_{i}^{s}$.

To capture these direct causal influences while accounting for uncertainty in physical processes, we model the relationship between a hidden variable and its parent nodes through a linear Gaussian structure:
\begin{equation}
z_{i}^{s} = \textbf{a}^T\mathcal{P}(z_{i}^{s}) + \sigma_i\boldsymbol{\epsilon}
\label{causal}
\end{equation}
\noindent where $\mathcal{P}(z_{i}^{s})$ is a vector of parent nodes of $z_{i}^{s}$, \textbf{a} represents the causal coefficients quantifying the strength of the influence of each parent, and $\boldsymbol{\epsilon} \sim N(0, I)$ introduces stochasticity to account for unmodeled factors and inherent randomness in physical processes.

While this causal structure captures the relationships between hidden variables, we also need to model how these variables relate to our background knowledge. The relationship between hidden variables and their background knowledge follows a log-normal distribution, which is commonly used in causal disaster systems~\cite{matgen2011towards, xu2022seismic}:
\begin{equation}
\log y_{i}^{k,s} = \theta_{0}^{i,k,s} + {\boldsymbol{\theta}_{\mathcal{P}(y_{i}^{s})}^{i,k,s}}^T\mathcal{P}(y_{i}^{s})(t) + \theta_{\epsilon}^{i,k,s}\boldsymbol{\epsilon}_i^{k,s}
\label{LogNormal}
\end{equation}
\noindent where $\boldsymbol{\epsilon}^{s} \sim N(0,I)$ and $\mathcal{P}(y_{i}^{s})$ represents the parent nodes of $y_{i}^{s}$. The dependency between $z_{i}^{s}$ and $y_{i}^{s}$ can be modeled by any reasonable distribution and depends on data characteristics and the underlying physical processes. In our causal Bayesian network, we model both physical variables and causal parameters that quantify these relationships as nodes.

Having established the static network relationships, we need to model how these variables evolve while preserving both causal and observational connections. Traditional static approaches cannot capture the dynamic nature of these relationships or the accumulation of uncertainty over time. Therefore, we employ an SDE framework that allows continuous evolution while maintaining our established relationships.

\noindent\textbf{Forward Diffusion Process.} 
For each hidden variable $z_{i}^{s}$, we define an SDE that incorporates both multi-scale background knowledge and causal dependencies:
\begin{equation}
\begin{aligned}\vspace{-0.3cm}
dz_{i}^{s}(t) = & \big\{f_{i}^{s}(z_{i}^{s}(t), \mathcal{P}(z_{i}^{s})(t), \textbf{a}^{s}, t)  \\
            & + \lambda_{i}^{s}(t)[\phi^{s}( y_{i}^{k,s}(t),\mathcal{P}(y_{i}^{s})(t), \boldsymbol{\theta}^{i,k,s}) - z_{i}^{s}(t)]\big\} dt + g_{i}^{s}(t)dW^{s}_{i}(t)
\end{aligned}
\label{forwardSDE}
\end{equation}
\noindent where $f_{i}^{s}(\cdot)$ and $g_{i}^{s}(\cdot)$ are neural networks capturing the deterministic and stochastic evolution respectively, while $\lambda_{i}^{s}(t)$ controls how strongly background knowledge influence the process.

The key to handling multi-resolution background knowledge lies in the $\phi$ function, which maps background knowledge back to the hidden state space defined in Equation \ref{LogNormal}:
\begin{equation}
\phi^s(y_{i}^{k,s}, \mathcal{P}(y_{i}^{s})(t), \boldsymbol{\theta}^{i,k,s}) = \frac{\log y_{i}^{k,s} - \theta_{0}^{i,k,s} - \boldsymbol{\theta}_{\mathcal{P}(y_{i}^{s}) \setminus z_{i}^{s}}^{i,k,s}(\mathcal{P}(y_{i}^{s}) \setminus z_{i}^{s})(t)}{\theta_{z_{i}^{s}}^{i,k,s}}
\label{phi_definition}
\end{equation}
\noindent where $\boldsymbol{\theta}^{i,k,s}$ is the resolution-specific parameters that allow this function to adapt its mapping based on the resolution of each background knowledge, effectively learning how information at different scales contributes to our understanding of the hidden state.

The initialization of our diffusion process differs between parent and child nodes:
For parent nodes, we use the finest available resolution background knowledge:
\begin{equation}
z_{i}^{s}(0) = \phi^s(y_{i}^{K_{i},s}, \mathcal{P}(y_{i}^{s})(0), \boldsymbol{\theta}^{i,K_i,s})
\label{origin}
\end{equation}

For child nodes, we balance both observational evidence and causal influences through a hybrid approach:
\begin{equation}
z_{i}^{s}(0) = a_{0}^{i,s} + (\textbf{a}_{\mathcal{P}(z_i^s)}^{i,s})^{T} \mathcal{P}(z_{i}^{s})(0) + a_{\phi}^{i,s}\cdot \phi^{s}(y_{i}^{k,s}, \mathcal{P}(y_{i}^{s})(0), \boldsymbol{\theta}^{i,k,s}) + a_{\epsilon}^{i,s}\cdot\boldsymbol{\epsilon}_{i}^{s}
\label{hybrid}
\end{equation}

This initialization framework consists of three interrelated components that together provide a comprehensive representation of initial state of the system. The causal component, represented by $(\textbf{a}_{\mathcal{P}(z_i^s)}^{i,s})^{T} \mathcal{P}(z_{i}^{s})$, captures the physical influence that parent nodes exert on their children.
The observational component, expressed as $a_{\phi}^{i,s}\cdot \phi^{s}(y_{i}^{k,s}, \mathcal{P}(y_{i}^{s})(0), \boldsymbol{\theta}^{i,k,s})$, grounds our model in empirical evidence by incorporating direct measurements. Through the $\phi$ function, this term skillfully handles background knowledge from different resolutions, transforming them into a consistent representation in the hidden variable space. This allows our model to leverage both high-resolution satellite imagery and coarser ground survey data within the same framework.
The stochastic component, given by $a_{\epsilon}^{i,s}\cdot\boldsymbol{\epsilon}_{i}^{s}$ where $\boldsymbol{\epsilon} \sim N(0, I)$, acknowledges and accounts for the inherent uncertainties present in both our measurements and our understanding of causal relationships. This term is crucial for capturing the natural variability in physical processes, measurement errors, and other unmodeled factors that influence our system. By including this stochastic element, our model can better represent the real-world uncertainty that characterizes complex disaster systems.

Beyond initialization, parent node influences persist throughout the diffusion process through the drift and diffusion terms. The neural networks $f_{i}^{s}(\cdot)$ and $g_{i}^{s}(\cdot)$ take both the current state and parent node states as inputs, enabling the model to capture non-linear, time-dependent causal relationships while maintaining computational feasibility.

\subsubsection{Semi-Implicit Variational Inference via Score Matching} \noindent\textbf{Reverse Diffusion Process.} The reverse diffusion process is used to recover the underlying distribution, effectively moving from low-resolution to high-resolution information. The reverse diffusion process is represented by a reverse SDE as defined in Equation \ref{ReverseSDE}.
\begin{equation}
\begin{aligned}
dz_{i}^{s}(t) &= \{f_{i}^{s}(z_{i}^{s}(t), \mathcal{P}(z_{i}^{s})(t), \textbf{a}^{s}, t) - g_{i}^{s}(t)^{2} \nabla_{z_{i}^{s}} \log p(z_{i}^{s}(t) \mid \mathcal{P}(z_{i}^{s})(t), \textbf{a}^{s}, \boldsymbol{\theta}^{s}) \\
&\quad + \lambda_{i}^{s}(t)[\phi^{s}( y_{i}^{k,s}(t),\mathcal{P}(y_{i}^{s})(t), \boldsymbol{\theta}^{i,k,s}) - z_{i}^{s}(t)]\}dt + g_{i}^{s}(t)d\bar{W}^{s}_{i}(t)
\end{aligned}
\label{ReverseSDE}
\end{equation}

\noindent where $p(z_{i}^{s}(t)\mid \mathcal{P}(z_{i}^{s})(t), \textbf{a}^{s}, \boldsymbol{\theta}^{s})$ is the conditional distribution of $z_{i}^{s}$ given its parents. $s_{i}^{s}(z_{i}^{s}(t), t, \textbf{a}^{s}, \boldsymbol{\theta}^{s}) = \nabla_{z_{i}^{s}} \log p(z_{i}^{s}(t)\mid \mathcal{P}(z_{i}^{s})(t), \textbf{a}^{s}, \boldsymbol{\theta}^{s})$ is the (conditional) score function.

\noindent\textbf{Score Matching Process} 
In practice, we first approximate the unconditional score for each variable with a neural network called the score net:
\vspace{-0.2cm}
\begin{equation}\vspace{-0.2cm}
 s_{\varphi}^{s}(z_{i}^{s}(t), t)  \approx \nabla_{z_{i}^{s}} \log p(z_{i}^{s}(t))
\label{SM_initial}
\end{equation}

We then design an objective to minimize a continuous weighted combination of Fisher divergences between $s_{\varphi}^{s}(z^{s}(t), t)$ and $\nabla\log p(z^{s}(t))$ through score matching \cite{vincent2011connection, song2020score}. To train each score net, we adapt the denoising score matching objective:
\vspace{-0.2cm}
\begin{equation}\vspace{-0.2cm}
     \mathcal{L}_{D} = \mathbb{E}_{t, z_{0}^{s}, \epsilon^{s}} \left[ \gamma^{s}(t) \psi^{s}(\sigma^{s}(t))\| s_{\varphi}^{s}(z_{t}^{s}, t) - \nabla_z \log p_{t|0}(z_{t}^{s} | z_{0}^{s}) \|^2 \right]
\label{FinalDLoss}
\end{equation}
\noindent where $p_{t|0}(z_{t}|z_{0}^{s})$ is the perturbation kernel representing the conditional probability density of state $z_{t}^{s}$ at time $t$, given initial state $z_{0}^{s}$ at time $0$. It describes how the initial distribution evolves over time according to the forward SDE. $\gamma^{s}(t)$ balances the importance of different time steps in the loss function. We use $\gamma^{s}(t) = \frac{1}{1+t}$, which gradually changes over time to ensure proper weighting across the entire trajectory. Other functions for $\gamma^{s}(t)$ can be chosen to suit specific modeling needs or empirical performance. $\psi^{s}(\sigma^{s}(t))=(\sigma^{s}(t))^2$ is chosen to ensure the score nets are trained to optimality~\cite{song2019generative}.

To compute the perturbation kernel $p_{t|0}(z_{t}^{s}|z_{0}^{s})$, we employ Bayes' theorem to set up relationships between conditional probabilities: \begin{equation}
p_{0|t}(z_{0}^{s} | z_{t}^{s}) = \frac{p_{t|0}(z_{t}^{s} | z_{0}^{s}) p_0(z_{0}^{s})}{p_t(z_{t}^{s})}
\end{equation}
Taking the logarithm and then the gradient with respect to $z_t^{s}$ yields:
\begin{equation}
\nabla_{z_t^{s}} \log p_{0|t}(z_0^{s} | z_t^{s}) = \nabla_{z_t^{s}} \log p_{t|0}(z_t^{s} | z_0^{s}) - \nabla_{z_t^{s}} \log p_t(z_t^{s})
\end{equation}
Note that $\nabla_{z_t^{s}} \log p_0(z_0^{s}) = 0$ as $z_0^{s}$ is independent of $z_t^{s}$. Rearranging terms, we obtain:
\begin{equation}
\nabla_{z_t} \log p_{t|0}(z_t^{s} | z_0^{s}) = \nabla_{z_t^{s}} \log p_t(z_t^{s}) + \nabla_{z_t^{s}} \log p_{0|t}(z_0^{s} | z_t^{s})
\label{Step5}
\end{equation}
For reversible processes satisfying detailed balance, the gradient of the log-probability in one direction is the negative of the gradient in the reverse direction. Consequently:
\begin{equation}
\nabla_{z_t^{s}} \log p_{t|0}(z_t^{s} | z_0^{s}) = - \nabla_{z_t^{s}} \log p_{0|t}(z_0^{s} | z_t^{s})
\label{perturbation}
\end{equation}
\noindent where the relationship between the perturbation kernel and the score function is crucial, as it enables us to train the score network using samples from the forward process.

In standard diffusion processes with simple SDEs (e.g., $d z = f(z,t)dt + g(t)dW$), the perturbation kernel often has a Gaussian expression: $p_{t|0}(z_t|z_0) = \mathcal{N}(z_t; \mu(t,z_0), \Sigma(t))$, where $\mu(t,z_0)$ and $\Sigma(t)$ are derived from the drift $f$ and diffusion $g$ terms. However, our case lacks an explicit perturbation kernel due to the drift and diffusion terms given by neural networks and the observation error term $\lambda_{i}^{s}(t)[\phi^{s}( y_{i}^{k,s}(t),\mathcal{P}(y_{i}^{s})(t), \boldsymbol{\theta}^{i,k,s}) - z_{i}^{s}(t)]$. To address this, we approximate the perturbation kernel $p_{t|0}(z_t^{s} | z_0^{s})$ using the Euler-Maruyama method (reparameterization):
\begin{equation}
z_{i}^{s}(t) \approx \mu_{i}^{s}(t, z_{i}(0)) + (\Sigma_{i}^{s})^{1/2}(t) \boldsymbol{\epsilon}_{i}^{s}
\label{EM}
\end{equation}
\noindent where $z_{i}^{s}(t)$ and $\mu_{i}^{s}(t, z_{i}^{s}(0))$ are d-dimensional vectors, $\mu_{i}^{s}(t, z_{i}(0))$ approximates the drift and observation terms, $\Sigma_{i}^{s}(t)$ is a d×d diagonal matrix approximating the diffusion term,
$\boldsymbol{\epsilon}_{i}^{s} \sim \mathcal{N}(\mathbf{0}, \mathbf{I})$ is a d-dimensional standard normal random vector. This reparameterization allows us to handle the complex structure of our model while maintaining computational feasibility.

We approximate $\mu^{s}(t, z_0^{s})$ and $\Sigma^{s}(t)$ in the forward process using the Euler-Maruyama method. Over the interval $[0, t]$, we have:
\begin{equation}
\begin{split}
z_{t}^{s} & \approx z_{0}^{s} + \int_{0}^{t} f_{i}^{s}(z_{i}^{s}(r), \mathcal{P}(z_i^{s})(r), \boldsymbol{a}, r)dr + \int_{0}^{t} g_{i}(r)dW_{i}(r) \\
& + \int_{0}^{t}  \lambda_{i}(r)[\phi( y_{i}^{k}(r),\mathcal{P}(y_{i})(r), \boldsymbol{\theta}^{i,k}) - z_{i}(r)]dr
\end{split}
\end{equation}

This approximates a Gaussian distribution:
\begin{equation}
p_{t|0}(z_t^{s}|z_0^{s}) \approx \mathcal{N}(z_t^{s}; \mu(t,z_0^{s}), \Sigma^{s}(t))
\label{ApproximateGaussian}
\end{equation}
\noindent where $\mu^{s}(t,z_0^{s})$ and $\Sigma^{s}(t)$ are defined as follows:
\begin{equation}
\begin{split}
\mu^{s}(t,z_0^{s}) & \approx z_0^{s} + \int_0^t f_i^{s}(z_i^{s}(r), \mathcal{P}(z_i^{s})(r), \boldsymbol{a}^{s}, r)dr \\
& + \int_0^t \lambda_{i}^{s}(r)[\phi^{s}( y_{i}^{k,s}(r),\mathcal{P}(y_{i}^{s})(r), \boldsymbol{\theta}^{i,k,s}) - z_{i}^{s}(r)]dr\\
\Sigma^{s}(t) & \approx \int_0^t \text{diag}(g_i^{s}(r)^2) dr
\end{split}
\label{mu_Sigma}
\end{equation}

Calculation of $\mu^{s}(t,z_0^{s})$ is performed using the Euler-Maruyama method. We divide the interval $[0, t]$ into $N$ steps of size $\Delta t = \frac{t}{N}$. Starting with $\mu_0^{s} = z_0^{s}$, we iteratively compute $\mu^{s}_j$ for $j = 1$ to $N$ using the equation:
\begin{equation}
\begin{split}
\mu_{j}^{s} & = \mu_{j-1}^{s} + [ f_i^{s}(\mu_{j-1}^{s}, \mathcal{P}(\mu_{j-1}^{s}), \boldsymbol{a}^{s}, (j-1)\Delta t) \\
& + \lambda_i^{s}((j-1)\Delta t) (\phi^{s}( y_{i}^{k,s}((j-1)\Delta t),\mathcal{P}(y_{i}^{s})((j-1)\Delta t), \boldsymbol{\theta}^{i,k,s}) - \mu_{j-1}^{s})] \Delta t
\end{split}
\label{EM_mu}
\end{equation}
\noindent where the final value, $\mu_N^{s}$, serves as our approximation of $\mu^{s}(T,z_0^{s})$.

For the calculation of $\Sigma^{s}(t)$, we employ Simpson's Rule with $N$ even steps. The approximation is given by:
\begin{equation}
\begin{aligned}
\Sigma^{s} & \approx \frac{\Delta t}{3} [ \text{diag}(g_{i}^{s}(0)^2) + 4 \sum_{j=1}^{N/2} \text{diag}(g_{i}^{s}((2j-1)\Delta t)^2) \\
&+ 2 \sum_{j=1}^{N/2-1} \text{diag}(g_i^{s}(2j\Delta t)^2) + \text{diag}(g_i^{s}(t)^2)]
\label{EM_Sigma}
\end{aligned}
\end{equation}

We begin with the score matching objective mentioned in Equation \ref{FinalDLoss}, where we approximate the perturbation kernel $p_{t|0}(z_t^{s} | z_0^{s})$ using the Euler-Maruyama method mention in Equation \ref{EM}. Taking the gradient with respect to $z_t^{s}$ and substituting $z_t^{s} - \mu^{s}(t, z_0^{s}) \approx (\Sigma^{s})^{1/2}(t) \boldsymbol{\epsilon}$, we obtain:
\(\nabla_{z_t^{s}} \log p(z_t^{s} | z_0^{s}) = - (\Sigma^{s})^{-1/2}(t) \boldsymbol{\epsilon}^{s} \).
By Equation \ref{perturbation}, we obtain:
\(\nabla_{z_{t}^{s}} \log p(z^{s}(0) | z^{s}(t)) = (\Sigma^{s})^{-1/2}(t) \boldsymbol{\epsilon}^{s} \).
This formulation allows us to estimate the score function using the reverse process. Our estimated score $s_{\varphi}^{s}$ should be:
\begin{equation}
\begin{aligned}
s_{\varphi}^{s}(z^{s}(t), t) & = \nabla_{z_{t}^{s}} \log p(z^{s}(t)|z^{s}(0))  \approx -\nabla{z_{t}^{s}} \log p(z^{s}(0)|z^{s}(t)) \\
& = -(\Sigma^{s})^{-1/2}(t) \boldsymbol{\epsilon}
\end{aligned}
\label{S_final}
\end{equation}
This leads to our final objective function:
\begin{equation}
\mathcal{L}_{D} = \mathbb{E}_{t, z_{0}^{s}, \boldsymbol{\epsilon}^{s}} \left[ \gamma^{s}(t) \psi^{s}(\Sigma^{s}(t)) || s_{\varphi}^{s}(z^{s}(t)) + (\Sigma^{s})^{-1/2}(t) \boldsymbol{\epsilon}^{s} ||^2 \right]
\label{L_SM}
\end{equation}
\noindent where $z^{s}(t) \approx \mu^{s}(t, z_{0}^{s}) + (\Sigma^{s})^{1/2}(t)\boldsymbol{\epsilon}^{s}$, $- (\Sigma^{s})^{-1/2}(t) \boldsymbol{\epsilon}^{s}$ is the target for $s^{s}_{\varphi}$, $\gamma^{s}(t)$ is a time-dependent weighting function, $\mu^{s}(t, z_0^{s})$ and $\Sigma^{s}(t)$ are derived from our specific drift and diffusion terms.

\noindent\textbf{Conditioning the Bayesian Network.}
The score matching process that have been discussed so far is the unconditional generative process. To incorporate the causal dependency in the generative process, we modify the score as in Equation \ref{SM_initial} with $\nabla\log p(z^{s}(t) | \mathcal{P}(z^{s}(t)))$ and plugging it back to the reverse SDE process.

To achieve this goal, several studies have proposed methods to approximate the conditional score using a single pre-trained score network, thus avoiding costly re-training~\cite{chung2022diffusion,song2020score,qu2024deep,rozet2023score}. These methods typically begin by expanding the conditional score using Bayes' rule, as shown in the following equation:
\begin{equation*}
\begin{split}
\nabla_{z^{s}(t)} \log p(z^{s}(t) | \mathcal{P}(z^{s})) & = \underbrace{\nabla \log p(z^{s}(t))}_{\text{unconditional score}} + \underbrace{\nabla\log p(\mathcal{P}(z^{s}) | z^{s}(t))}_{\text{causal score}} 
\end{split}
\end{equation*}

\noindent where the first term on the right-hand side is approximated by the pre-trained score function. The primary challenge lies in accurately estimating the second term, which we refer to as the causal score. This decomposition allows us to leverage our pre-trained score network for the prior score and combine it with the likelihood score to infer various observation scenarios without retraining.

Given the causal structure defined in Equation \ref{causal}, where $z_i^{s}$ is a child node with parent nodes $\mathcal{P}(z_i^{s})$, the conditional distribution of $\mathcal{P}(z_i^{s})$ given $z_i^{s}$ follows a multivariate normal distribution. This choice is theoretically well-founded as complex systems often exhibit local Gaussian behavior due to the interaction of multiple underlying processes \cite{friedrich1997description,gardiner2004handbook}, particularly through the effects of the generalized Central Limit Theorem on interacting components.
\vspace{-0.2cm}
\begin{equation}\vspace{-0.2cm}
    \mathcal{P}(z_i^{s}) | z_i^{s} \sim \mathcal{N}(\mu_c^{s}(z_i^{s}), \Sigma_c^{s})
\label{parentConditionalDist}
\end{equation}
\noindent where the conditional mean vector \(\mu_c^{s}(z_i^{s})\) is computed by adjusting the base vector with respect to \(z_i^{s}\): \(\mu_c^{s}(z_i^{s}) = \mu_{\mathcal{P}(z_i^{s})} + \Sigma^{s}_{\mathcal{P}(z_i^{s}),z_i^{s}} (\Sigma^{s}_{z_i^{s}})^{-1}(z_i^{s} - \mu_i^{s})\). The conditional covariance matrix \(\Sigma_c^{s}\) integrates the modifications: \(\Sigma_c^{s}=\Sigma_{\mathcal{P}(z_i)}^{s}-\Sigma_{\mathcal{P}(z_i),z_i}^{s} (\Sigma_{z_i^{s}}^{s})^{-1}\Sigma_{\mathcal{P}(z_i^{s}),z_i^{s}}^{s}\). This formulation is powerful for complex systems as: (1) the conditional mean $\mu_c(z_i)$ naturally incorporates causal influences \cite{pearl2009,spirtes2001causation}, (2) the covariance structure $\Sigma_c$ systematically captures uncertainty propagation crucial in geophysical systems \cite{assimilation2009ensemble}, and (3) enables efficient computation while maintaining theoretical guarantees about uncertainty propagation \cite{sarkka2013spatiotemporal}.

Based on the Gaussian assumption in Equation~\ref{parentConditionalDist}, the conditional probability distribution \( p(\mathcal{P}(z_i^{s}) | z_i^{s}(t)) \) can be approximated as follows~\cite{chung2022diffusion}:
\vspace{-0.2cm}
\begin{equation}
p(\mathcal{P}(z_i^{s}) | z_i^{s}(t)) = \int p(\mathcal{P}(z_i^{s}) | z_i^{s})p(z_i^{s} | z_i^{s}(t)) \,dz_i^{s} \simeq \mathcal{N}(\mu_c^{s}(\hat{z}_i^{s}(z_i^{s}(t))), \Sigma_c^{s}),
\label{likelihoondApprox}
\end{equation}
\vspace{-0.2cm}
where \( \mu_c^{s} \) and \( \Sigma_c^{s} \) are determined by the specific formulation of the model. The mean \( \hat{z}_i^{s}(z_{i}^{s}(t)) \) is estimated using Tweedie's formula~\cite{kim2021noise2score}, which provides a closed-form approximation for the posterior expectation. Specifically, the mean \( \hat{z}_i^{s}(z_{i}^{s}(t)) \) is given by:
\vspace{-0.2cm}
\begin{equation}
\hat{z}_i^{s}(z_{i}^{s}(t)) = \mathbb{E}_{p(z_i^{s}|z_i^{s}(t))}[z_i^{s}] \simeq \frac{z_i^{s}(t) + \Sigma_{i}^{s}(t) s_{\phi}^{s}(z_i^{s}(t),t)}{\mu_i^{s}(t)},
\end{equation}
\vspace{-0.2cm}
where \( \Sigma_i^{s}(t) \) and \( \mu_i^{s}(t) \) represent the covariance and mean derived from the numerical method in the forward diffusion process, and \( s_{\phi}^{s}(z_i^{s}(t), t) \) is the score function parameterized by the model.

Then the conditional score can be calculated as:
\begin{equation}
\begin{split}
 \nabla_{z_i^{s}(t)} \log p(z_i^{s}(t) | \mathcal{P}(z_i^{s})) & = \nabla_{z_i^{s}(t)} \log p(z_i^{s}(t)) \\
 & + \nabla_{z_i^{s}(t)} \log p(\mathcal{P}(z_i^{s}) | z_i^{s}(t))
\label{ConditonalScore}
\end{split}
\end{equation}

The decomposition of the conditional score into unconditional and causal components is theoretically justified by the structure of complex systems. As demonstrated in fluid dynamics \cite{friedrich1997description} and general stochastic systems \cite{gardiner2004handbook}, local Gaussian approximations often provide accurate descriptions of system behavior even when the global dynamics are highly non-linear. This locality principle allows us to effectively capture causal dependencies while maintaining computational tractability.

The first term of RHS in Equation \ref{ConditonalScore} is given by the score network and the second gradient is calculated by chain rule and automatic differentiation (auto-grad) according to the relationship discussed above:
\begin{equation}
\begin{split}
& \nabla_{z_i^{s}(t)} \log p(\mathcal{P}(z_i^{s}) | z_{i}^{s}(t)) = \frac{\partial \log \mathcal{N}(\mu_c^{s}(\hat{z}_i^{s}(z_i^{s}(t))), \Sigma_c^{s})}{\partial z_i^{s}(t)} \\
&= \frac{\partial \log \mathcal{N}(\mu_c^{s}(\hat{z}_i^{s}(z_i^{s}(t))), \Sigma_c^{s})}{\partial \mu_c^{s}} \frac{\partial \mu_c^{s}}{\partial \hat{z}_i^{s}} \frac{\partial \hat{z}_i^{s}}{\partial z_i^{s}(t)} 
\end{split}
\label{S_condition}
\end{equation}
\noindent where the division of the last line is element-wise.

This approach of decomposing the conditional score into unconditional and causal components allows us to leverage pre-trained score networks while incorporating causal dependencies. It offers a computationally efficient way to handle complex causal structures without the need for retraining the entire model for each new conditioning scenario. The use of Tweedie's formula in estimating $\hat{z}_i^{s}(z_i^{s}(t))$ further enhances the efficiency of the method by avoiding the need to compute complex integrals explicitly.

\textbf{Numerical Stability Improvement.} While the decomposition above provides a theoretical foundation for incorporating causal dependencies, practical implementation faces numerical challenges. In particular, the approximation in Equation \ref{likelihoondApprox} assumes perfect estimation of $\hat{z}_{i}^{s}(z_{i}^{s}(t))$, which becomes increasingly unreliable at high noise levels. The noise comes from the non-negligible variance given by \(p(z|z(t))\). 

The challenge is further complicated by the non-linear nature of our forward process, where the drift and diffusion coefficients are learned through neural networks. This results in a non-linear conditional distribution $p(z_{i}^{s}(t)|z_{i}^{s})=\mathcal{N}(\mu^{s}(z_{i}^{s},t), \Sigma^{s}(t))$. To derive a stable approximation, we take inspiration from analyzing local linearization of such systems.

To motivate our solution, consider a simplified case where $x \sim \mathcal{N}(\mu_x, \Sigma_x)$ and the forward process $x(t)|x \sim \mathcal{N}(\mu(x,t), \Sigma(t))$. Let $\hat{x} = \mathbb{E}_{p(x|x(t))}[x]$ denote the conditional mean of $x$ given $x(t)$, similar to $\hat{z}_i(z_i(t))$ defined earlier. By performing first-order Taylor expansion of $\mu(x,t)$ around $\hat{x}$:
\begin{equation}
   \mu(x,t) \approx \mu(\hat{x},t) + J_{\hat{x}}(x - \hat{x})
\end{equation}
where $J_{\hat{x}} = \partial\mu(x,t)/\partial x|_{x=\hat{x}}$. Under this local linearization, we can derive the conditional distribution:
\begin{equation}
   p(x|x(t)) \approx \mathcal{N}(\mu_{x|x(t)}, \Sigma_{x|x(t)})
\end{equation}
where
\begin{align}
   \mu_{x|x(t)} &= \mu_x + \Sigma_xJ_{\hat{x}}^T(J_{\hat{x}}\Sigma_xJ_{\hat{x}}^T + \Sigma(t))^{-1}(x(t) - \mu(\hat{x},t) - J_{\hat{x}}(\mu_x - \hat{x})) \\
   \Sigma_{x|x(t)} &= \Sigma_x - \Sigma_xJ_{\hat{x}}^T(J_{\hat{x}}\Sigma_xJ_{\hat{x}}^T + \Sigma(t))^{-1}J_{\hat{x}}\Sigma_x
\end{align}

Applying this analysis to our causal setting, we propose a modified approximation for the parent node distribution:
\begin{equation}
    p(\mathcal{P}(z_{i}^{s})|z_{i}^{s}(t)) \approx \mathcal{N}(\mu_{c}^{s}(\hat{z}_{i}^{s}(z_{i}^{s}(t))), \Sigma_{c}^{s} + \Sigma^{s}_{z_{i}^{s}|z_{i}^{s}(t)})
\end{equation}
where $\mu_{c}^{s}$ and $\Sigma_{c}^{s}$ follow the formulation in Equation \ref{parentConditionalDist}, and $\Sigma^{s}_{z_{i}^{s}|z_{i}^{s}(t)}$ takes the form:
\begin{equation}
    \Sigma^{s}_{z_i^{s}|z_i^{s}(t)} = \Sigma^{s}_{z_i^{s}} - \Sigma^{s}_{z_i^{s}}J_{\hat{z}_i^{s}}^T(J_{\hat{z}_i^{s}}\Sigma^{s}_{z_i^{s}}J_{\hat{z}_i^{s}}^T + \Sigma_i^{s}(t))^{-1}J_{\hat{z}_i^{s}}\Sigma^{s}_{z_i^{s}}
\end{equation}

This modification provides an automatic mechanism for gradient stabilization: at high noise levels, the increased uncertainty naturally reduces gradient magnitudes, preventing unstable updates, while at low noise levels, the uncertainty reduces to maintain precise estimation. This approach offers a principled way to balance between stability and accuracy across different noise regimes.

To ensure robust sampling, we employ Langevin Monte Carlo in the simulation:
\begin{equation}
z^{s}(t) \leftarrow z^{s}(t) + \kappa^{s}_{\varphi^{s}}(z^{s}(t), t) + \sqrt{2\kappa^{s}} (\boldsymbol{\Sigma}^{s})^{1/2} \boldsymbol{\epsilon}^{s}
\end{equation}

\noindent\textbf{Stochastic Variational Inference.}
Building on this formulation, we develop an inference algorithm that jointly estimates probability distributions of unobserved target variables and their causal dependencies. We employ variational inference to transform the inference problem into an optimization task.
Our goal is to jointly infer the true posteriors of multiple unobserved target variables $z_i^{s}$, along with unknown causal dependency parameters. We use variational inference to approximate the true posteriors of $Z^{s}$ using a variational distribution $q(Z^{s})$. The optimization objective function combines variational lower bounds to best approximate the true posterior distribution. Specifically, we optimize the ELBO to find optimal parameters for $q(Z^{s})$ that closely match the true posterior introduced in Section \ref{VIIntro}.

We derive the lower bound of the marginal log-likelihood of the observed $Y^{s}$ by Jensen’s inequality as follows~\cite{jordan1998introduction}:
\begin{equation}
\begin{split}
\log p(Y^{s}) & \geq \mathcal{L}_{v} =\mathbb{E}_{q^{s}}[ \underbrace{\log p(Y^{s}|Z^{s})}_{[1]} + \underbrace{\log p(Z^{s})}_{[2]} - \underbrace{\log q(Z^{s}|Y^{s})}_{[3]}]
\label{ELBO}
\end{split}
\end{equation}
\noindent where $q^{s}$ is obtained from the reverse diffusion process and has a complex, unknown distribution. Consequently, direct derivation of the variational bound is intractable. This challenge necessitates the development of alternative approaches to optimize the ELBO and perform inference effectively.

We expand the likelihood term $[1]$ in Equation \ref{ELBO} based on the assumption that background knowledge follow log-normal distributions as stated in Equation \ref{LogNormal}, with the distribution of their parent nodes formulated in Equation \ref{EM}. Therefore, we can get the distribution $y_{i}^{k_i,s}|\mathcal{P}(y_i^{s}) \sim LN(\mathbf{C}_{i,k_{i}}^{s}, \mathbf{D}_{i,k_{i}}^{s})$, we have:
\begin{align}
\mathbf{C}_{i,k_{i}}^{s} & = \theta_{0}^{i,k_{i},s} + {\boldsymbol{\theta}_{\mathcal{P}(y_{i})}^{i,k_{i},s}}^T\mu_{\mathcal{P}(y_{i}^{s})}^{s}(t_{k_{i}}, z^{s}_{\mathcal{P}(y_{i}^{s})}(t_{k_i}))\\
&\mathbf{D}_{i,k_{i}}^{s} = {\boldsymbol{\theta}_{\mathcal{P}(y_{i})}^{i,k_{i},s}}\Sigma^{s}_{\mathcal{P}(y_{i}^{s})}(t_{k_i}){\boldsymbol{\theta}_{\mathcal{P}(y_{i}^{s})}^{i,k_{i},s}}^T + (\eta_{i,k_{i}}^{s})^{2}\mathbf{I}
\end{align}.  
\noindent where $\mu^{s}$ and $\Sigma^{s}$ are derived in Equation \ref{mu_Sigma}.

The formulation of term $[1]$ is then presented as follows:
\begin{align*}
\log p(Y^{s}|Z^{s}) 
& = \sum_{i,k_{i}} \big\{-\frac{1}{2} (\log y_i^{k_{i},s} - \mathbf{C}_{i,k_{i}}^{s})^T (\mathbf{D}_{i,k_{i}}^{s})^{-1} (\log y_i^{k_{i},s} - \mathbf{C}_{i,k_{i}}^{s}) \\
& - \frac{1}{2} \log(2\pi) - \frac{1}{2} \log |\mathbf{D}_{i,k_{i}}^{s}| - \sum_{d_{i}=1} \log y_{i,d_{i}}^{k_{i},s} \big\}
\label{likelihood}
\end{align*}
\noindent where $d_{i}$ represents the element index of the vector $y_{i}^{k_{i},s}$.

Given the Gaussian prior distributions assumption and causal relationship assumption, the term [2] in Equation \ref{ELBO} can be derived as $\log p(Z^{s}) = \sum_{i}\log p(z_i^{s}|\mathcal{P}(z_{i}^{s}))$, where $p(z_i^{s}|\mathcal{P}(z_{i}^{s})) = \mathcal{N}(z_i^{s}|\mu^{s}_{z_i|\mathcal{P}}, \Sigma^{s}_{z_i|\mathcal{P}})$.
To calculate the item [3] in Equation \ref{ELBO}, we approximate the posterior for each unknown variable $z^{s}_{i}$ through the reverse diffusion process. To achieve this, we start with $z_T^{s}$ given by a forward diffusion path and run the reverse SDE from $t = T$ to $t = 0$ as in Equation \ref{ReverseSDE}. The final step from the reverse diffusion process is a sample from $q(z_i^{s}|y^{s})$. We run the reverse process to get $M$ samples of $z_i^{s}(0)$, allowing us to get $M$ entire reverse sequences. 

Consequently, the M reconstructed samples $z_i^{s}(0)$ form a complex posterior distribution that approximates the unknown variable $z_i^{s}$, incorporating multi-resolution background knowledge and causal dependency. Eventually, we estimate $\log q_(z_i^{s}|y^{s})$, apply kernel density estimation (KDE):
\begin{equation}
\log q_(z_i^{s}|y^{s}) \approx \log \left( \frac{1}{Mh^D} \sum_{l=1}^M K \left( \frac{z_i^{s} - z^{l,s}_i}{h} \right) \right)
\end{equation}
\noindent where \( K(u) = \frac{1}{(2\pi)^{D/2}} \exp\left(-\frac{1}{2} u^T u\right) \) is a Gaussian kernel and \(h\) is the bandwidth which can be determined by Scott's rule. Besides, the expectation of the ELBO in Equation \ref{ELBO} is approximated by averaging the linearly combined terms [1], [2], and [3] over M samples of \( z_i(0) \).

The ELBO is defined and decomposed as follows: 


\begin{equation}
\begin{split}
\mathcal{L}_{v} 
& = \frac{1}{LM} \sum_{l} \Bigg\{\sum_{m,i,j} \log p(y_{i}^{j,l,s}|\mathcal{P}(y_i^{m,l,s})) + \sum_{m,i}\log p(z_{i}^{m,l,s}|\mathcal{P}(z_{i}^{m,l,s})) - \sum_{m,i} \log q(z_{i}^{m,l,s}|y^{l,s})\Bigg\}\\
& = \frac{1}{LM} \sum_{l} \Bigg\{\sum_{m,i,j} \Big[-\frac{1}{2} (\log y_i^{l,s} - \mathbf{C}_{m,i,l}^{s})^T (\mathbf{D}_{m,i,l}^{s})^{-1} (\log y_i^{l,s} - \mathbf{C}_{m,i,l}^{s}) - \frac{1}{2} \log(2\pi) - \frac{1}{2} \log |\mathbf{D}_{m,i,l}^{s}| - \sum_{d_{i}=1} \log y_{i,d_{i}}^{l,s} \Big] \\
& - \sum_{m,i}  \Big[\frac{1}{2} \log(2\pi) + \frac{1}{2} (z_i^{m,l,s}-\mu_{z_i^{s}|\mathcal{P}^{s}})^T(\Sigma_{z_i|\mathcal{P}}^{s})^{-1}(z_i^{m,l,s}-\mu^{s}_{z_i^{s}|\mathcal{P}}) + \frac{1}{2} \log |\Sigma^{s}_{z_i^{s}|\mathcal{P}}|\Big] \\
& - \sum_{m,i} \log \left( \frac{1}{Mh^D} \sum_{k=1}^M K \left( \frac{z_i^{m,l,s} - z^{k,l,s}_i}{h} \right) \right) \Bigg\}
\label{finalELBO}
\end{split}
\end{equation}

We optimize our framework by minimizing a combined loss from Equations \ref{FinalDLoss} and \ref{finalELBO}, seeking optimal posteriors and parameters that best explain the observed data while respecting the causal structure.

\noindent \textbf{Optimization and Computational Efficiency.} To enhance computational efficiency and scalability, especially when dealing with large-scale datasets or long sequences typical in Earth system processes, we implement mini-batch processing with stochastic gradient descent (SGD). Instead of using the entire dataset for each update, we randomly sample a subset of the data for each iteration. This approach not only reduces memory requirements but also allows for faster iterations and potentially better generalization. This approach also facilitates the use of stochastic optimization techniques, which can lead to faster convergence and better escape from local optima. We update weights at iteration $t + 1$ as:
\(\textbf{w}_{t+1} = \textbf{w}_{t} + \rho A \nabla_{\textbf{w}}\mathcal{L}\), where $\textbf{w}$ includes all trainable parameters, $A$ is a positive definite preconditioner \cite{ji2020variational, paisley2011discrete}, $\rho$ is the learning rate, and $\nabla_{\textbf{w}}\mathcal{L}$ are loss function gradients. This update scheme converges to a local optimum if $\rho$ meets conditions in \cite{robbins1951stochastic}. 

\section{Code Availability}\label{Appendix_code}
The source codes for the implementation of the multi-resolution variational diffusion framework for modeling
complex systems with causal dependency are publicly available online under accession code: https://github.com/PaperSubmissionFinal/KDD2025.

\section{Data Availability}\label{Appendix_data}

\noindent\textbf{Multi-resolution Earthquake Dataset:} The multi-resolution datasets utilized in this study are derived from publicly accessible sources. The primary data sources include high-resolution satellite imagery in the form of Damage Proxy Maps (DPMs), which were produced by NASA's Advanced Rapid Imaging and Analysis (ARIA) team using Copernicus Sentinel-1 satellite data. These maps are accessible through the ARIA Data Share platform \url{https://aria-share.jpl.nasa.gov/}.

Additional data sources include ground failure models and USGS ShakeMap data, specifically the Peak Ground Acceleration (PGA) measurements used to generate building fragility curves. These can be obtained from the United States Geological Survey's Earthquake Hazards Program website \url{https://earthquake.usgs.gov/}.

\noindent\textbf{Multi-resolution Fire Spread Dataset:} For the fire spread prediction task, we utilize a diverse set of multi-resolution data sources obtained through Google Earth Engine \cite{gerard2023wildfirespreadts}. At the foundation of our dataset is the SRTM Digital Elevation Model, providing high-resolution topographical data at 30m resolution, which is crucial for understanding terrain effects on fire behavior.

The VIIRS satellite system provides several key data products: active fire detection at 375m resolution with 24-hour temporal frequency, surface reflectance measurements at varying resolutions between 500m and 1km, and vegetation indices at 500m resolution updated every 8 days. These data are complemented by MODIS-derived land cover classification at 500m resolution, providing essential information about fuel types and vegetation patterns.

Weather and climate conditions, critical factors in fire spread, are captured through two main sources. The GRIDMET dataset provides detailed meteorological variables including wind, rain, temperature, humidity, and drought indices at approximately 4.6km resolution with daily updates. This is supplemented by NOAA GFS weather forecast data at approximately 28km resolution with hourly predictions, enabling both current condition assessment and near-term forecasting capabilities. This diverse range of data sources enables a comprehensive analysis of multi-resolution dynamics and causal dependencies in fire spread behavior.

\printbibliography 

@article{rozet2023score,
  title={Score-based data assimilation},
  author={Rozet, Fran{\c{c}}ois and Louppe, Gilles},
  journal={Advances in Neural Information Processing Systems},
  volume={36},
  pages={40521--40541},
  year={2023}
}

@article{patten2024data,
  title={Data-Driven Earthquake Multi-impact Modeling: A Comparison of Models},
  author={Patten, Hamish and Anderson Loake, Max and Steinsaltz, David},
  journal={International Journal of Disaster Risk Science},
  pages={1--13},
  year={2024},
  publisher={Springer}
}

@article{song2021maximum,
  title={Maximum likelihood training of score-based diffusion models},
  author={Song, Yang and Durkan, Conor and Murray, Iain and Ermon, Stefano},
  journal={Advances in neural information processing systems},
  volume={34},
  pages={1415--1428},
  year={2021}
}

@article{song2020improved,
  title={Improved techniques for training score-based generative models},
  author={Song, Yang and Ermon, Stefano},
  journal={Advances in neural information processing systems},
  volume={33},
  pages={12438--12448},
  year={2020}
}

@inproceedings{sohl2015deep,
  title={Deep unsupervised learning using nonequilibrium thermodynamics},
  author={Sohl-Dickstein, Jascha and Weiss, Eric and Maheswaranathan, Niru and Ganguli, Surya},
  booktitle={International conference on machine learning},
  pages={2256--2265},
  year={2015},
  organization={PMLR}
}

@article{saharia2022image,
  title={Image super-resolution via iterative refinement},
  author={Saharia, Chitwan and Ho, Jonathan and Chan, William and Salimans, Tim and Fleet, David J and Norouzi, Mohammad},
  journal={IEEE transactions on pattern analysis and machine intelligence},
  volume={45},
  number={4},
  pages={4713--4726},
  year={2022},
  publisher={IEEE}
}

@article{dhariwal2021diffusion,
  title={Diffusion models beat gans on image synthesis},
  author={Dhariwal, Prafulla and Nichol, Alexander},
  journal={Advances in neural information processing systems},
  volume={34},
  pages={8780--8794},
  year={2021}
}

@article{kingma2021variational,
  title={Variational diffusion models},
  author={Kingma, Diederik and Salimans, Tim and Poole, Ben and Ho, Jonathan},
  journal={Advances in neural information processing systems},
  volume={34},
  pages={21696--21707},
  year={2021}
}

@article{ho2020denoising,
  title={Denoising diffusion probabilistic models},
  author={Ho, Jonathan and Jain, Ajay and Abbeel, Pieter},
  journal={Advances in neural information processing systems},
  volume={33},
  pages={6840--6851},
  year={2020}
}

@inproceedings{arjovsky2017wasserstein,
  title={Wasserstein generative adversarial networks},
  author={Arjovsky, Martin and Chintala, Soumith and Bottou, L{\'e}on},
  booktitle={International conference on machine learning},
  pages={214--223},
  year={2017},
  organization={PMLR}
}

@inproceedings{qu2024deep,
  title={Deep generative data assimilation in multimodal setting},
  author={Qu, Yongquan and Nathaniel, Juan and Li, Shuolin and Gentine, Pierre},
  booktitle={Proceedings of the IEEE/CVF Conference on Computer Vision and Pattern Recognition},
  pages={449--459},
  year={2024}
}

@article{vincent2011connection,
  title={A connection between score matching and denoising autoencoders},
  author={Vincent, Pascal},
  journal={Neural computation},
  volume={23},
  number={7},
  pages={1661--1674},
  year={2011},
  publisher={MIT Press}
}

@article{brunton2016discovering,
  title={Discovering governing equations from data by sparse identification of nonlinear dynamical systems},
  author={Brunton, Steven L and Proctor, Joshua L and Kutz, J Nathan},
  journal={Proceedings of the national academy of sciences},
  volume={113},
  number={15},
  pages={3932--3937},
  year={2016},
  publisher={National Acad Sciences}
}

@article{torres2021deep,
  title={Deep learning for time series forecasting: a survey},
  author={Torres, Jos{\'e} F and Hadjout, Dalil and Sebaa, Abderrazak and Mart{\'\i}nez-{\'A}lvarez, Francisco and Troncoso, Alicia},
  journal={Big Data},
  volume={9},
  number={1},
  pages={3--21},
  year={2021}
}

@article{vaswani2017attention,
  title={Attention is all you need},
  author={Vaswani, A},
  journal={Advances in Neural Information Processing Systems},
  year={2017}
}

@article{song2019generative,
  title={Generative modeling by estimating gradients of the data distribution},
  author={Song, Yang and Ermon, Stefano},
  journal={Advances in neural information processing systems},
  volume={32},
  year={2019}
}

@article{kim2021noise2score,
  title={Noise2score: tweedie’s approach to self-supervised image denoising without clean images},
  author={Kim, Kwanyoung and Ye, Jong Chul},
  journal={Advances in Neural Information Processing Systems},
  volume={34},
  pages={864--874},
  year={2021}
}

@inproceedings{de2011multi,
  title={Multi-scale assimilation of AMSR-E snow water equivalent and MODIS snow cover fraction in northern Colorado},
  author={De Lannoy, GJ and Reichle, RH and Arsenault, KR and Houser, PR and Kumar, Sujay and Verhoest, Niko and Pauwels, VR},
  booktitle={AGU Fall Meeting Abstracts},
  volume={2011},
  pages={H13I--03},
  year={2011}
}

@article{reichstein2019deep,
  title={Deep learning and process understanding for data-driven Earth system science},
  author={Reichstein, Markus and Camps-Valls, Gustau and Stevens, Bjorn and Jung, Martin and Denzler, Joachim and Carvalhais, Nuno and Prabhat, F},
  journal={Nature},
  volume={566},
  number={7743},
  pages={195--204},
  year={2019},
  publisher={Nature Publishing Group UK London}
}

@article{song2019survey,
  title={A survey of remote sensing image classification based on CNNs},
  author={Song, Jia and Gao, Shaohua and Zhu, Yunqiang and Ma, Chenyan},
  journal={Big earth data},
  volume={3},
  number={3},
  pages={232--254},
  year={2019},
  publisher={Taylor \& Francis}
}

@article{atkinson2013downscaling,
  title={Downscaling in remote sensing},
  author={Atkinson, Peter M},
  journal={International Journal of Applied Earth Observation and Geoinformation},
  volume={22},
  pages={106--114},
  year={2013},
  publisher={Elsevier}
}

@inproceedings{xu2014assimilation,
  title={Assimilation of SMOS soil moisture in the MESH model with the ensemble Kalman filter},
  author={Xu, Xiaoyong and Li, Jonathan and Tolson, Bryan A and Staebler, Ralf M and Seglenieks, Frank and Davison, Bruce and Haghnegahdar, Amin and Soulis, Eric D},
  booktitle={2014 IEEE Geoscience and Remote Sensing Symposium},
  pages={3766--3769},
  year={2014},
  organization={IEEE}
}

@article{matgen2011towards,
  title={Towards an automated SAR-based flood monitoring system: Lessons learned from two case studies},
  author={Matgen, P and Hostache, R and Schumann, G and Pfister, L and Hoffmann, L and Savenije, HHG},
  journal={Physics and Chemistry of the Earth, Parts A/B/C},
  volume={36},
  number={7-8},
  pages={241--252},
  year={2011},
  publisher={Elsevier}
}

@article{cutter2015global,
  title={Global risks: Pool knowledge to stem losses from disasters},
  author={Cutter, Susan L and Ismail-Zadeh, Alik and Alc{\'a}ntara-Ayala, Irasema and Altan, Orhan and Baker, Daniel N and Brice{\~n}o, Salvano and Gupta, Harsh and Holloway, Ailsa and Johnston, David and McBean, Gordon A and others},
  journal={Nature},
  volume={522},
  number={7556},
  pages={277--279},
  year={2015},
  publisher={Nature Publishing Group UK London}
}

@article{shi2015convolutional,
  title={Convolutional LSTM network: A machine learning approach for precipitation nowcasting},
  author={Shi, Xingjian and Chen, Zhourong and Wang, Hao and Yeung, Dit-Yan and Wong, Wai-Kin and Woo, Wang-chun},
  journal={Advances in neural information processing systems},
  volume={28},
  year={2015}
}

@article{gerard2023wildfirespreadts,
  title={WildfireSpreadTS: A dataset of multi-modal time series for wildfire spread prediction},
  author={Gerard, Sebastian and Zhao, Yu and Sullivan, Josephine},
  journal={Advances in Neural Information Processing Systems},
  volume={36},
  pages={74515--74529},
  year={2023}
}

@article{li2024rapid,
  title={Rapid Building Damage Estimates From the M7. 8 Turkey Earthquake Sequence via Causality-Informed Bayesian Inference From Satellite Imagery},
  author={Li, Xuechun and Yu, Xiao and B{\"u}rgi, Paula M and Wald, David J and Hu, Xie and Xu, Susu},
  journal={Earthquake Spectra},
  pages={87552930241290501},
  year={2024},
  publisher={SAGE Publications Sage UK: London, England}
}

@article{chen2018neural,
  title={Neural ordinary differential equations},
  author={Chen, Ricky TQ and Rubanova, Yulia and Bettencourt, Jesse and Duvenaud, David K},
  journal={Advances in neural information processing systems},
  volume={31},
  year={2018}
}

@article{nowicki2018global,
  title={A global empirical model for near-real-time assessment of seismically induced landslides},
  author={Nowicki Jessee, MA and Hamburger, MW and Allstadt, K and Wald, David J and Robeson, SM and Tanyas, H and Hearne, M and Thompson, EM},
  journal={Journal of Geophysical Research: Earth Surface},
  volume={123},
  number={8},
  pages={1835--1859},
  year={2018},
  publisher={Wiley Online Library}
}

@article{zhu2015geospatial,
  title={A geospatial liquefaction model for rapid response and loss estimation},
  author={Zhu, Jing and Daley, Davene and Baise, Laurie G and Thompson, Eric M and Wald, David J and Knudsen, Keith L},
  journal={Earthquake Spectra},
  volume={31},
  number={3},
  pages={1813--1837},
  year={2015},
  publisher={GeoScienceWorld}
}

@article{song2020score,
  title={Score-based generative modeling through stochastic differential equations},
  author={Song, Yang and Sohl-Dickstein, Jascha and Kingma, Diederik P and Kumar, Abhishek and Ermon, Stefano and Poole, Ben},
  journal={arXiv preprint arXiv:2011.13456},
  year={2020}
}

@inproceedings{li2023disasternet,
  title={DisasterNet: Causal Bayesian networks with normalizing flows for cascading hazards estimation from satellite imagery},
  author={Li, Xuechun and B{\"u}rgi, Paula M and Ma, Wei and Noh, Hae Young and Wald, David Jay and Xu, Susu},
  booktitle={Proceedings of the 29th ACM SIGKDD Conference on Knowledge Discovery and Data Mining},
  pages={4391--4403},
  year={2023}
}

@article{wang2024scalable,
  title={Scalable and rapid building damage detection after hurricane Ian using causal Bayesian networks and InSAR imagery},
  author={Wang, Chenguang and Liu, Yepeng and Zhang, Xiaojian and Li, Xuechun and Paramygin, Vladimir and Sheng, Peter and Zhao, Xilei and Xu, Susu},
  journal={International Journal of Disaster Risk Reduction},
  pages={104371},
  year={2024},
  publisher={Elsevier}
}

@inproceedings{wang2023causality,
  title={Causality-informed Rapid Post-hurricane Building Damage Detection in Large Scale from InSAR Imagery},
  author={Wang, Chenguang and Liu, Yepeng and Zhang, Xiaojian and Li, Xuechun and Paramygin, Vladimir and Subgranon, Arthriya and Sheng, Peter and Zhao, Xilei and Xu, Susu},
  booktitle={Proceedings of the 8th ACM SIGSPATIAL International Workshop on Security Response using GIS},
  pages={7--12},
  year={2023}
}

@inproceedings{li2024optimizing,
  title={Optimizing Rapid Seismic Building Damage Assessment: Integrating Enhanced Radar Change Detection Maps with Variational Bayesian Networks},
  author={Li, Xuechun and Gao, Runyu and Burgi, Paula M and Wald, David J and Xu, Susu},
  booktitle={IGARSS 2024-2024 IEEE International Geoscience and Remote Sensing Symposium},
  pages={3791--3795},
  year={2024},
  organization={IEEE}
}

@article{li2024scalable,
  title={Scalable Variational Learning for Noisy-OR Bayesian Networks with Normalizing Flows for Complex Cascading Disaster Systems},
  author={Li, Xuechun and Xu, Susu},
  year={2024}
}

@article{li2024spatial,
  title={Spatial-variant causal Bayesian inference for rapid seismic ground failures and impacts estimation},
  author={Li, Xuechun and Xu, Susu},
  journal={arXiv preprint arXiv:2412.00026},
  year={2024}
}

@article{li2025multi,
  title={Multi-class Seismic Building Damage Assessment from InSAR Imagery using Quadratic Variational Causal Bayesian Inference},
  author={Li, Xuechun and Xu, Susu},
  journal={arXiv preprint arXiv:2502.18546},
  year={2025}
}

@article{li2025rapid,
  title={Rapid building damage estimates from the M7. 8 Turkey earthquake sequence via causality-informed Bayesian inference from Satellite Imagery},
  author={Li, Xuechun and Yu, Xiao and B{\"u}rgi, Paula M and Wald, David J and Hu, Xie and Xu, Susu},
  journal={Earthquake Spectra},
  volume={41},
  number={1},
  pages={5--33},
  year={2025},
  publisher={SAGE Publications Sage UK: London, England}
}

@article{li2023normalizing,
  title={Normalizing flow-based deep variational bayesian network for seismic multi-hazards and impacts estimation from insar imagery},
  author={Li, Xuechun and Burgi, Paula M and Ma, Wei and Noh, Hae Young and Wald, David J and Xu, Susu},
  journal={arXiv preprint arXiv:2310.13805},
  year={2023}
}

@article{li2024multi,
  title={Multi-resolution Data Assimilation for Cascading Hazard Modeling: A Causality-informed Multi-scale Stochastic Differential Equation Framework},
  author={Li, Xuechun and Gao, Shan and Gao, Runyu and Xu, Susu},
  journal={AGU24},
  year={2024},
  publisher={AGU}
}

@inproceedings{xu2022deep,
  title={Deep Causal Bayesian Network for Modeling Spatial Seismic Building Damage from Remote Sensing Observations},
  author={Xu, Susu and Li, Xuechun and Noh, Haeyoung and Wald, David J},
  booktitle={AGU Fall Meeting Abstracts},
  volume={2022},
  pages={INV44A--02},
  year={2022}
}

@article{xu2022seismic,
  title={Seismic multi-hazard and impact estimation via causal inference from satellite imagery},
  author={Xu, Susu and Dimasaka, Joshua and Wald, David J and Noh, Hae Young},
  journal={Nature Communications},
  volume={13},
  number={1},
  pages={7793},
  year={2022},
  publisher={Nature Publishing Group UK London}
}

@article{raissi2019physics,
  title={Physics-informed neural networks: A deep learning framework for solving forward and inverse problems involving nonlinear partial differential equations},
  author={Raissi, Maziar and Perdikaris, Paris and Karniadakis, George E},
  journal={Journal of Computational physics},
  volume={378},
  pages={686--707},
  year={2019},
  publisher={Elsevier}
}

@inproceedings{ranganath2014black,
  title={Black box variational inference},
  author={Ranganath, Rajesh and Gerrish, Sean and Blei, David},
  booktitle={Artificial intelligence and statistics},
  pages={814--822},
  year={2014},
  organization={PMLR}
}

@article{hoffman2014no,
  title={The No-U-Turn sampler: adaptively setting path lengths in Hamiltonian Monte Carlo.},
  author={Hoffman, Matthew D and Gelman, Andrew and others},
  journal={J. Mach. Learn. Res.},
  volume={15},
  number={1},
  pages={1593--1623},
  year={2014}
}

@article{kucukelbir2017automatic,
  title={Automatic differentiation variational inference},
  author={Kucukelbir, Alp and Tran, Dustin and Ranganath, Rajesh and Gelman, Andrew and Blei, David M},
  journal={Journal of machine learning research},
  volume={18},
  number={14},
  pages={1--45},
  year={2017}
}

@inproceedings{yin2018semi,
  title={Semi-implicit variational inference},
  author={Yin, Mingzhang and Zhou, Mingyuan},
  booktitle={International conference on machine learning},
  pages={5660--5669},
  year={2018},
  organization={PMLR}
}

@article{yang2023diffusion,
  title={Diffusion models: A comprehensive survey of methods and applications},
  author={Yang, Ling and Zhang, Zhilong and Song, Yang and Hong, Shenda and Xu, Runsheng and Zhao, Yue and Zhang, Wentao and Cui, Bin and Yang, Ming-Hsuan},
  journal={ACM Computing Surveys},
  volume={56},
  number={4},
  pages={1--39},
  year={2023},
  publisher={ACM New York, NY, USA}
}

@article{yu2017spatio,
  title={Spatio-temporal graph convolutional networks: A deep learning framework for traffic forecasting},
  author={Yu, Bing and Yin, Haoteng and Zhu, Zhanxing},
  journal={arXiv preprint arXiv:1709.04875},
  year={2017}
}

@article{palmer2019scientific,
  title={The scientific challenge of understanding and estimating climate change},
  author={Palmer, Tim and Stevens, Bjorn},
  journal={Proceedings of the National Academy of Sciences},
  volume={116},
  number={49},
  pages={24390--24395},
  year={2019},
  publisher={National Acad Sciences}
}

@book{pearl2009,
author = {Pearl, Judea},
title = {Causality: Models, Reasoning and Inference},
year = {2009},
isbn = {052189560X},
publisher = {Cambridge University Press},
address = {USA},
edition = {2nd}
}

@article{jordan1998introduction,
  title={An introduction to variational methods for graphical models},
  author={Jordan, Michael I and Ghahramani, Zoubin and Jaakkola, Tommi S and Saul, Lawrence K},
  journal={Learning in graphical models},
  pages={105--161},
  year={1998},
  publisher={Springer}
}

@article{robbins1951stochastic,
  title={A stochastic approximation method},
  author={Robbins, Herbert and Monro, Sutton},
  journal={The annals of mathematical statistics},
  pages={400--407},
  year={1951},
  publisher={JSTOR}
}

@inproceedings{paisley2011discrete,
  title={The discrete infinite logistic normal distribution for mixed-membership modeling},
  author={Paisley, John and Wang, Chong and Blei, David},
  booktitle={Proceedings of the Fourteenth International Conference on Artificial Intelligence and Statistics},
  pages={74--82},
  year={2011},
  organization={JMLR Workshop and Conference Proceedings}
}

@inproceedings{ji2020variational,
  title={Variational training for large-scale noisy-OR Bayesian networks},
  author={Ji, Geng and Cheng, Dehua and Ning, Huazhong and Yuan, Changhe and Zhou, Hanning and Xiong, Liang and Sudderth, Erik B},
  booktitle={Uncertainty in Artificial Intelligence},
  pages={873--882},
  year={2020},
  organization={PMLR}
}

@book{gardiner2004handbook,
  title={Handbook of Stochastic Methods for Physics, Chemistry and the Natural Sciences},
  author={Gardiner, Crispin W},
  year={2004},
  publisher={Springer}
}

@article{friedrich1997description,
  title={Description of a turbulent cascade by a Fokker-Planck equation},
  author={Friedrich, Rudolf and Peinke, Joachim},
  journal={Physical review letters},
  volume={78},
  number={5},
  pages={863},
  year={1997},
  publisher={APS}
}

@article{sarkka2013spatiotemporal,
  title={Spatiotemporal learning via infinite-dimensional Bayesian filtering and smoothing: A look at Gaussian process regression through Kalman filtering},
  author={S{\"a}rkk{\"a}, Simo and Solin, Arno and Hartikainen, Jouni},
  journal={IEEE Signal Processing Magazine},
  volume={30},
  number={4},
  pages={51--61},
  year={2013},
  publisher={IEEE}
}

@article{assimilation2009ensemble,
  title={The Ensemble Kalman Filter},
  author={Assimilation, Data},
  journal={IEEE CONTROL SYSTEMS MAGAZINE},
  volume={1066},
  number={033X/09},
  year={2009}
}

@book{spirtes2001causation,
  title={Causation, prediction, and search},
  author={Spirtes, Peter and Glymour, Clark and Scheines, Richard},
  year={2001},
  publisher={MIT press}
}

@article{chung2022diffusion,
  title={Diffusion posterior sampling for general noisy inverse problems},
  author={Chung, Hyungjin and Kim, Jeongsol and Mccann, Michael T and Klasky, Marc L and Ye, Jong Chul},
  journal={arXiv preprint arXiv:2209.14687},
  year={2022}
}

\section*{Acknowledgement}
The author(s) disclosed receipt of the following financial support for the research, authorship, and/ or publication of this article: X. L. and S. X. are supported by U.S. Geological Survey Grant G22AP00032 and NSF CMMI-2242590.

\section*{Author contributions}
X.L., S.G., S.X. conceptualized the research, developed the Temporal-SVGDM framework. X.L. processed the experimental data. X.L. and S.G. implemented the code, conducted the experiments, analyzed the results. All authors edited the paper and approved the submission.

\section*{Competing interests}
The authors declare no competing interests.

\newpage
\section{Supplementary information}

\subsection{Building damage estimation for the 2021 Haiti earthquake}

\begin{figure*}[ht]
  \centering
\includegraphics[width=1.1\linewidth]{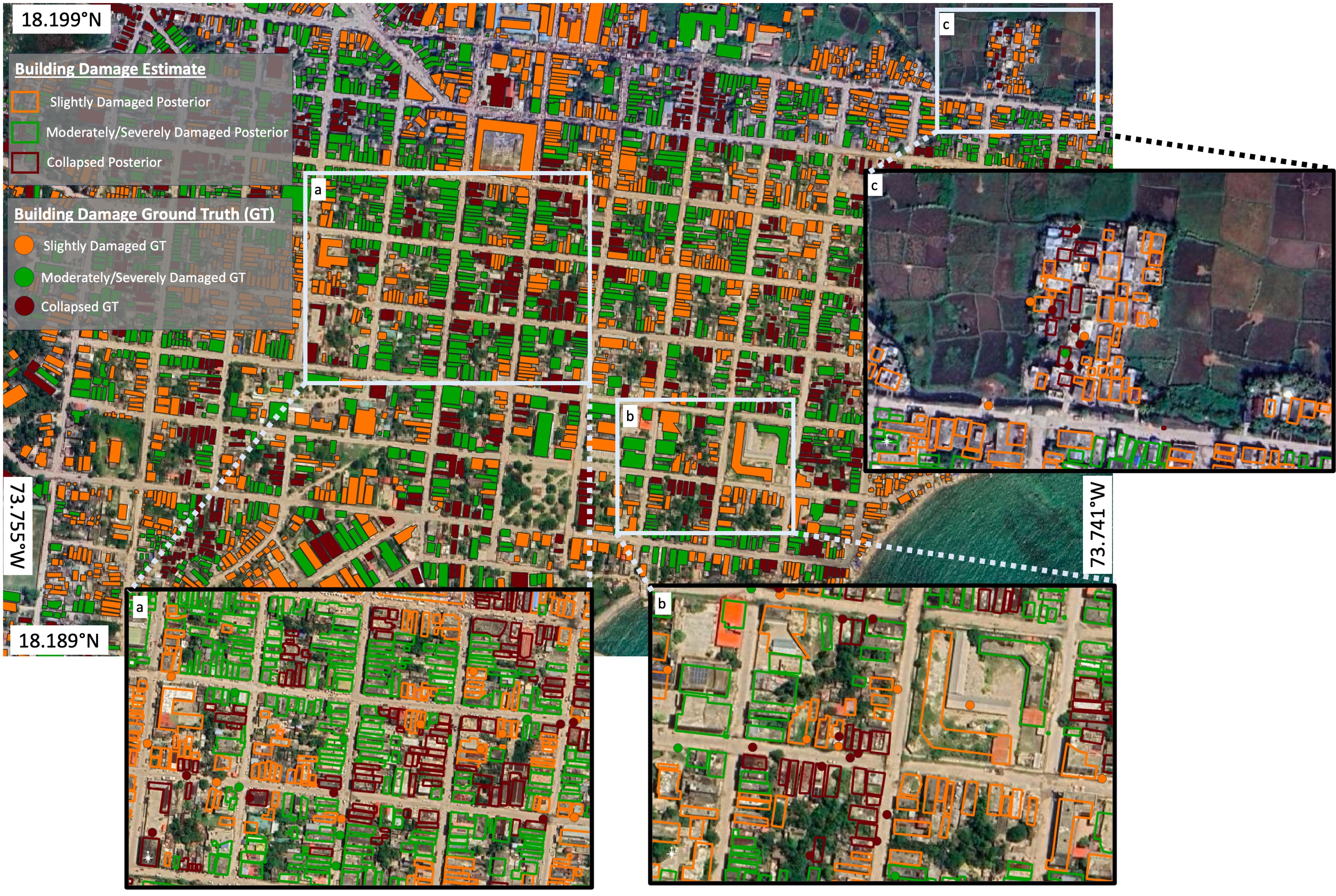}
  \caption{\textbf{Estimated building damage probability maps with ground truth lavel for the 2021 Haiti earthquake study.} Figures (a)–(c) present the building damage estimates with ground truth information. The legend colors represent the
building damage levels, as indicated.}
  \label{Haiti_BD}
    \end{figure*}

Figure \ref{Haiti_BD} shows the building damage estimation in Haiti's densely populated urban areas, demonstrating the ability of our model to distinguish damage states between adjacent buildings while handling complex architectural layouts.

\newpage

\subsection{Building damage assessment for the 2020 Puerto Rico earthquake}
\begin{figure*}[ht]
  \centering
\includegraphics[width=1.1\linewidth]{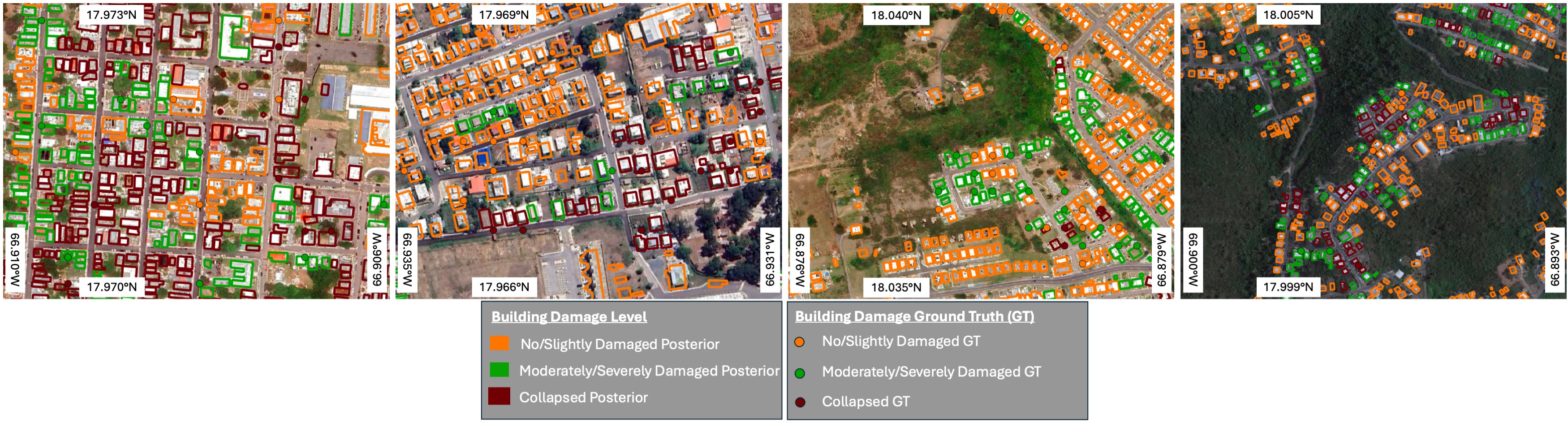}
  \caption{Estimated building damage probability maps with ground truth lavel for the 2020 Puerto Rico earthquake study.}
  \label{PR_BD}
    \end{figure*}

Figure \ref{PR_BD} presents the estimated building damage probabilities demonstrate strong agreement with ground truth observations across varying damage levels from slight damage to complete collapse in urban areas of Puerto Rico.

\subsection{Estimated liquefaction probability for the 2020 Puerto Rico earthquake}

\begin{figure*}[ht]
  \centering
\includegraphics[width=1.1\linewidth]{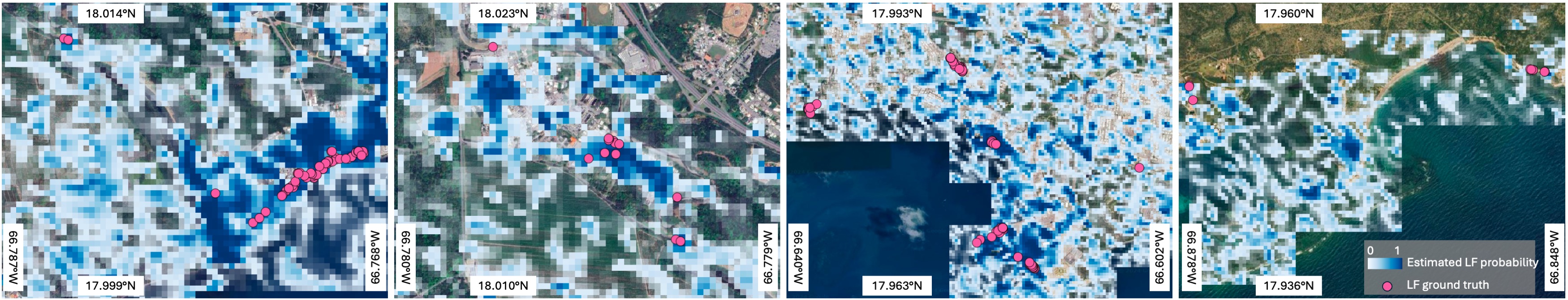}
\caption{Visualization of estimated liquefaction probability maps with ground truth labels in the 2020 Puerto Rico earthquake case.}
  \label{PR_LF}
    \end{figure*}

The visualization of estimated liquefaction probability maps, presented in Figure \ref{PR_LF}, shows a close correlation with the observations of ground truth in different geological settings, from coastal areas to inland regions.

\newpage

\subsection{Active fire prediction visualization}

\begin{figure*}[ht]
  \centering
\includegraphics[width=1\linewidth]{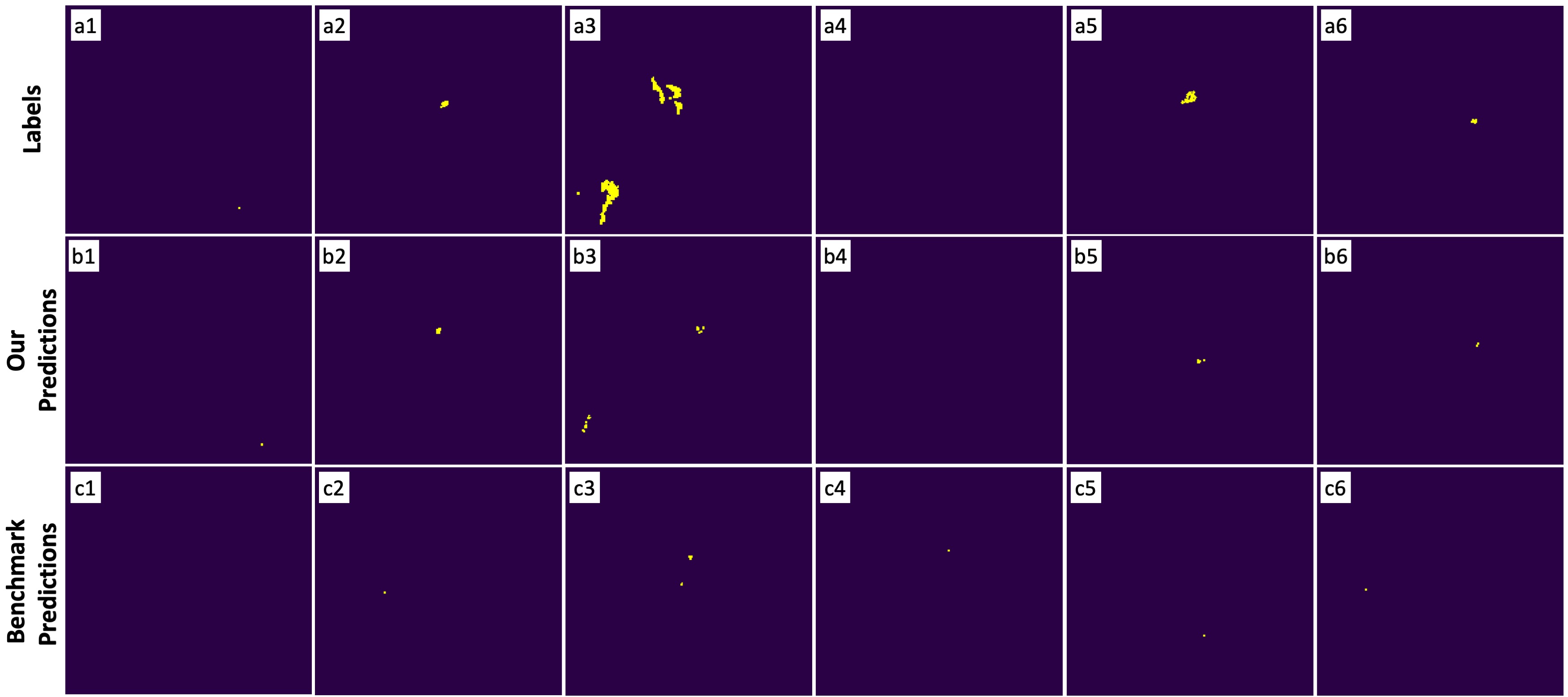}
  \caption{\textbf{Active fire detection and prediction results.} Comparison of fire detection performance across six consecutive time steps (columns 1-6). Top row (a1-a6): Ground truth fire locations. Middle row (b1-b6): Temporal-SVGDM predictions. Bottom row (c1-c6): Benchmark model predictions. Yellow points indicate detected fire locations, demonstrating the model's ability to capture both individual fire occurrences and fire clusters while maintaining temporal consistency.}
  \label{ActiveFire}
    \end{figure*}

The comparison between model predictions and ground truth observations demonstrates Temporal-SVGDM's capability in capturing wildfire dynamics. Figure \ref{ActiveFire} presents a comprehensive comparison across six time steps, showing ground truth labels (top row), Temporal-SVGDM predictions (middle row), and benchmark model predictions (bottom row). This visualization highlights our model's ability to accurately detect fire locations while maintaining both spatial precision and temporal consistency in fire progression patterns.

\end{document}